\documentclass{article}

\usepackage{microtype}
\usepackage{graphicx}
\usepackage{subcaption}
\usepackage{booktabs} 
\usepackage{enumitem}
\usepackage{tabularx}

\usepackage{algorithm}
\usepackage{algpseudocode}

\usepackage[most]{tcolorbox}  
\usepackage{xcolor}
\usepackage{colortbl}
\usepackage{siunitx}

\usepackage{multirow} 
\usepackage{float}

\usepackage{hyperref}

\usepackage[accepted]{icml2025}
\makeatletter
\@ifundefined{ICML@appearing}{%
    \newcommand{\ICML@appearing}{}%
}{}
\makeatother

\usepackage{amsmath}
\usepackage{amssymb}
\usepackage{mathtools}
\usepackage{amsthm}
\usepackage{makecell} 
\usepackage{enumitem}

\usepackage[capitalize,noabbrev]{cleveref}

\theoremstyle{plain}

\theoremstyle{definition}

\theoremstyle{remark}

\usepackage[textsize=tiny]{todonotes}

\newcounter{promptcnt}  

\definecolor{qaBlue}{RGB}{25,55,95}        
\definecolor{searchTeal}{RGB}{0,105,92}    
\definecolor{solveAmber}{RGB}{191,144,0}   
\definecolor{criticRed}{RGB}{183,28,28}    
\definecolor{selectPurple}{RGB}{81,45,168} 
\definecolor{refineIndigo}{RGB}{40,53,147} 

\definecolor{midnight}{RGB}{25,55,95}      
\definecolor{iceblue}{RGB}{245,248,252}    

\newtcolorbox{promptbox}[3][]{
    enhanced,
    colback=iceblue,
    colframe=midnight,
    fonttitle=\bfseries\sffamily\color{white},
    title={#2~\refstepcounter{promptcnt}},
    label={#3},  
    attach boxed title to top left={xshift=5mm,yshift=-3mm},
    boxed title style={
        colback=midnight,
        sharp corners,
        boxrule=0pt,
        left=4mm,right=4mm,
        top=1mm,bottom=1mm
    },
    sharp corners,
    left=5mm,right=5mm,top=9mm,bottom=5mm,
    boxrule=0.6pt,
    shadow={2mm}{-2mm}{0mm}{black!20},
    breakable,
    #1
}

\icmltitlerunning{ReThinker: Scientific Reasoning by Rethinking with Guided Reflection and Confidence Control}

\begin{document}

\twocolumn[
\icmltitle{ReThinker: Scientific Reasoning by Rethinking with Guided Reflection and Confidence Control}



\icmlsetsymbol{equal}{*}

\begin{icmlauthorlist}
\icmlauthor{Zhentao Tang}{equal,noah}
\icmlauthor{Yuqi Cui}{equal,noah}
\icmlauthor{Shixiong Kai}{noah}
\icmlauthor{Wenqian Zhao}{noah}
\icmlauthor{Ke Ye}{noah}
\icmlauthor{Xing Li}{noah}
\icmlauthor{Anxin Tian}{noah}
\icmlauthor{Zehua Pei}{cuhk}
\icmlauthor{Hui-Ling Zhen}{noah}
\icmlauthor{Shoubo Hu}{noah}
\icmlauthor{Xiaoguang Li}{noah}
\icmlauthor{Yunhe Wang}{noah}
\icmlauthor{Mingxuan Yuan}{noah}
\end{icmlauthorlist}

\icmlaffiliation{noah}{Noah's Ark Lab, Huawei, China}
\icmlaffiliation{cuhk}{The Chinese University of Hong Kong, Hong Kong, China}

\icmlcorrespondingauthor{Shixiong Kai}{kaishixiong@huawei.com}

\icmlkeywords{Machine Learning}

\vskip 0.3in
]



\printAffiliationsAndNotice{\icmlEqualContribution} 

\begin{abstract}
Expert-level scientific reasoning remains challenging for large language models, particularly on benchmarks such as Humanity’s Last Exam (HLE), where rigid tool pipelines, brittle multi-agent coordination, and inefficient test-time scaling often limit performance.
We introduce ReThinker, an confidence-aware agentic framework that orchestrates retrieval, tool use, and multi-agent reasoning through a stage-wise Solver--Critic--Selector architecture. Rather than following a fixed pipeline, ReThinker dynamically allocates computation based on model confidence, enabling adaptive tool invocation, guided multi-dimensional reflection, and robust confidence-weighted selection.
To support scalable training without human annotation, we further propose a reverse data synthesis pipeline and an adaptive trajectory recycling strategy that transform successful reasoning traces into high-quality supervision. 
Experiments on HLE, GAIA, and XBench demonstrate that ReThinker consistently outperforms state-of-the-art foundation models with tools and existing deep research systems, achieving state-of-the-art results on expert-level reasoning tasks.
\end{abstract}

\section{Introduction}

\begin{figure}[t]
    \centering 
    \includegraphics[width=0.95\linewidth]{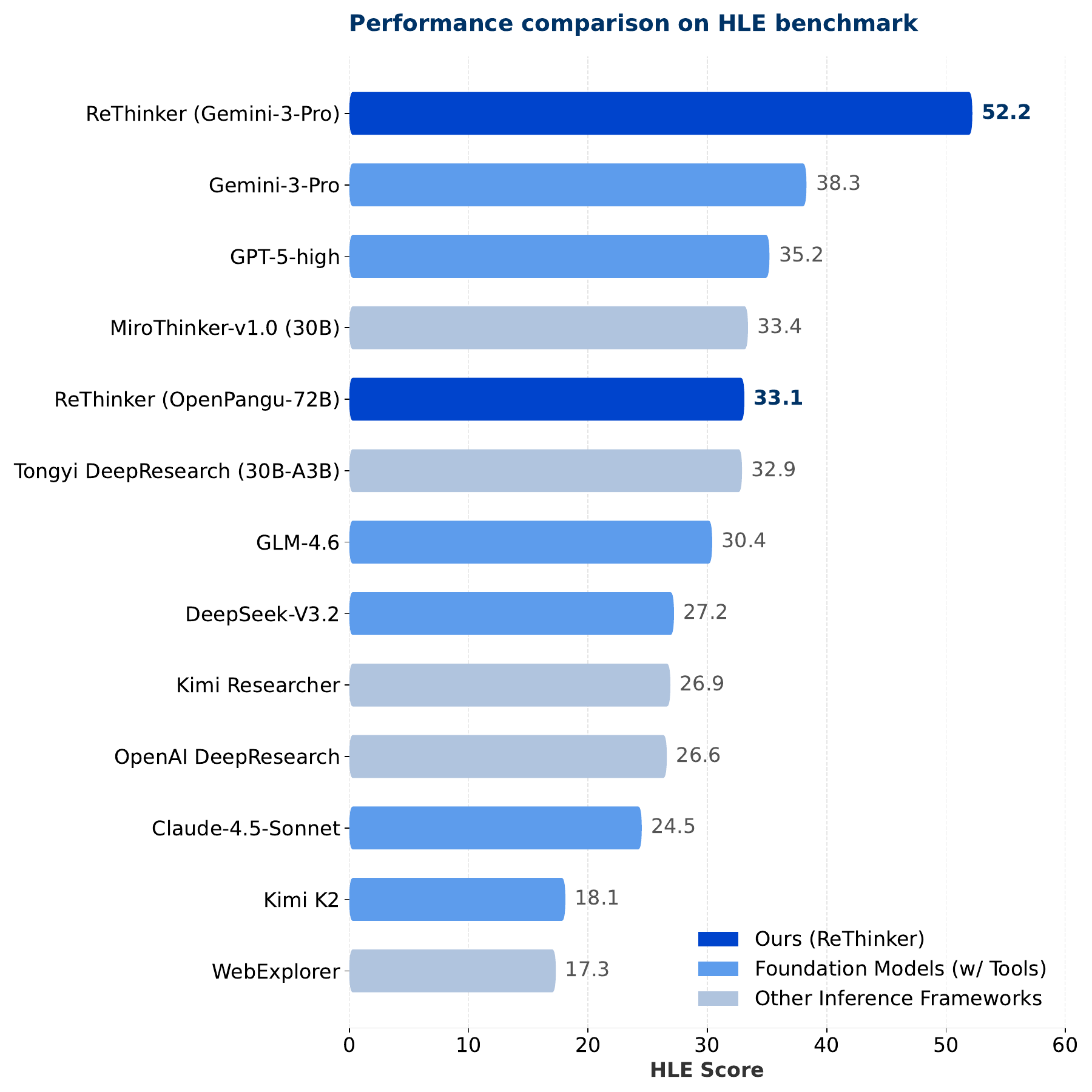}
    \caption{Performance comparison on the HLE benchmark. The results include Foundation Models with Tools, existing Inference Frameworks, and our proposed method ReThinker based on two LLMs. ReThinker (based on Gemini-3-Pro) significantly outperforming both standalone models and other inference frameworks.}
    \label{fig:fig_intro}
\end{figure}

Scientific reasoning has become a central challenge for evaluating the capabilities of large language models (LLMs) and a key indicator of progress toward general-purpose artificial intelligence~\cite{truhn2023large}. In contrast to commonsense reasoning, scientific problem-solving demands quantitative rigor, multi-hop causal inference, and the integration of domain-specific knowledge across mathematics, physics, and chemistry—capabilities that remain insufficiently developed in current LLMs. This limitation becomes particularly evident on expert-level benchmarks such as Humanity’s Last Exam (HLE)~\cite{phan2025humanity}, which targets advanced scientific problems requiring deep domain expertise and complex multi-step reasoning. Although existing LLMs often exhbit strong superficial performance, they frequently fail to reliably distinguish correct mathematical reasoning from subtly flawed arguments, suggesting that their apparent success is driven more by pattern memorization than by systematic, principled deduction.


To address these limitations, we argue that expert-level scientific reasoning demands three fundamental capabilities that remain critically underdeveloped in current systems: the capacity for \textbf{rethinking}—iteratively questioning and refining intermediate conclusions rather than committing to single-pass reasoning trajectories; the mechanism for \textbf{guided reflection}—structured, dimension-specific error diagnosis that transcends superficial summarization to target precise logical, strategic, and knowledge gaps; and effective  \textbf{confidence control}—explicit uncertainty quantification and multi-round adjudication to stabilize answer selection amidst compounding verification noise. 
Here we introduce \textbf{ReThinker}. Our contributions are summarized as follows:

\begin{itemize}

\item \textbf{Automated Trajectory Synthesis for Rethinking Supervision.} We eliminate manual annotation entirely. Our system automatically generates expert-level QA pairs across scientific domains. It extracts domain concepts from web contexts and generated trajectories. The pipeline records complete multi-stage reasoning traces. It captures error recovery patterns and tool-use sequences. Only verified correct trajectories are retained. These provide high-fidelity supervision signals. Models learn to rethink rather than memorize patterns.

\item \textbf{Hybrid Scaling with Guided Reflection.} We develop a hybrid sequential–parallel scaling architecture based on EvoFabric~\cite{evofabric2025} that enables flexible trade-offs between inference budget and reasoning accuracy. The framework integrates Python execution, web search, and web parsing tools to support quantitative verification and expert knowledge acquisition. In the \textbf{Solver} stage, we employ multi-round iterative synthesis to allow progressive refinement of reasoning. In the \textbf{Critic} stage, we introduce a summary-and-guidance module that processes the complete prior trajectory, mitigating context-length limitations and correcting subtle errors that are often missed by conventional summary-only critics.

\item \textbf{Confidence-Controlled Selection via Uncertainty Aggregation.} We introduce a confidence-guided multi-round selection mechanism for the \textbf{Selector} stage to stabilize optimal answer identification. To address verification-induced uncertainty, we aggregate perplexity-based internal consistency metrics across multiple selection rounds. Prior selection outcomes and confidence scores are iteratively fed back into the prompt to amplify high-confidence candidates. To further eliminate ordering bias, we permute candidate positions using Latin Square designs and resolve cross-round inconsistencies through a final adjudication step to determine the definitive answer.

\end{itemize}

\section{Related Work}

\subsection{Tool-Augmented Interactive Reasoning}
The ReAct framework~\cite{yao2022react} turns LLMs into interactive agents by interleaving \textit{Thought--Action--Observation} steps, enabling tool use during reasoning. In scientific settings, ReAct-style agents employ calculators for symbolic computation~\cite{chen2022program}, code execution for mathematical and logical verification~\cite{wang2024executable,m2024augmenting}, and web search for evidence and literature retrieval~\cite{nakano2021webgpt}. Recent extensions such as \textbf{Eigen-1}~\cite{tang2025eigen} further integrate reasoning with executable Python-based tool workflows and report strong performance on HLE Bio/Chem Gold. \textbf{SCOPE}~\cite{pei2025scope} automates prompt evolution to improve agent effectiveness, reducing reliance on manual prompt engineering. However, most tool-augmented approaches remain largely single-agent: reasoning depth is constrained by context length, errors can accumulate without systematic correction, and tool hallucination remains a persistent challenge~\cite{zhang2025siren}, motivating multi-agent decomposition.

\subsection{Multi-Agent Orchestration and Collaborative Reasoning}
Multi-agent systems decompose complex reasoning tasks into specialized roles that collaborate through parallel or sequential interaction patterns~\cite{xi2025rise}. 
Complementary to role-based coordination, recent work on \textbf{self-reflection}, such as MiroThinker~\cite{team2025mirothinker}, shows that agents trained on trajectories containing explicit error-correction steps can achieve improved reasoning performance. The \textbf{STeP} method~\cite{chen2025training} further synthesizes self-reflective trajectories from teacher models, enabling smaller open-source models to acquire corrective behaviors. Despite these advances, most existing multi-agent and self-reflective frameworks rely on hand-crafted interaction protocols and do not explicitly model confidence or answer stability under test-time scaling, limiting their robustness on challenging reasoning benchmarks.

\subsection{Test-Time Scaling and Confidence-Guided Reflection}
Test-time scaling (also referred to as inference-time scaling) has emerged as an effective strategy for enhancing reasoning performance without model retraining~\cite{yang2023large}. Existing approaches broadly fall into two categories.

\textbf{Sequential scaling} extends reasoning trajectories through iterative reflection and revision. For example, \textbf{s1}~\cite{muennighoff2025s1} demonstrates that budget forcing produces longer and more accurate reasoning traces, while \textbf{Reflexion}~\cite{shinn2023reflexion} incorporates verbal feedback stored in episodic memory to guide subsequent reasoning steps. Despite improved reasoning depth, sequential methods remain limited by single-trajectory exploration and accumulated errors.

\textbf{Parallel scaling} generates multiple candidate solutions simultaneously and selects the optimal answer through verification or aggregation~\cite{snell2025scaling}. Representative approaches include \textbf{Best-of-N} sampling~\cite{ichihara2025evaluation}, which ranks candidates using reward models or process verifiers~\cite{lightman2023let}, and \textbf{self-consistency}~\cite{wang2023self}, which performs majority voting across diverse reasoning paths. However, verifier-based methods introduce substantial computational overhead and often rely on auxiliary models~\cite{zhengprobabilistic}.

The effectiveness of parallel scaling depends critically on accurate confidence estimation. Existing approaches can be broadly categorized into two classes. \textbf{Consistency-based methods}~\cite{zhou2025theoretical} measure agreement across multiple samples, with self-consistency as a representative example. While effective for deterministic problems, such metrics can be unstable when reasoning paths diverge or verification signals are noisy~\cite{chen2024universal}. \textbf{Probability-based methods} leverage internal model statistics, with \textbf{perplexity} commonly used as a confidence indicator~\cite{chen1999empirical}. Recent theoretical analyses~\cite{murugadoss2025evaluating} suggest that perplexity correlates with reasoning path quality; however, single-round confidence estimates remain unreliable due to ordering bias and sampling variance~\cite{bito2025evaluating}.

\section{Method}
\subsection{Overall Framework Overview}

\begin{figure*}[!t]
\centering
\includegraphics[width=0.8\linewidth]{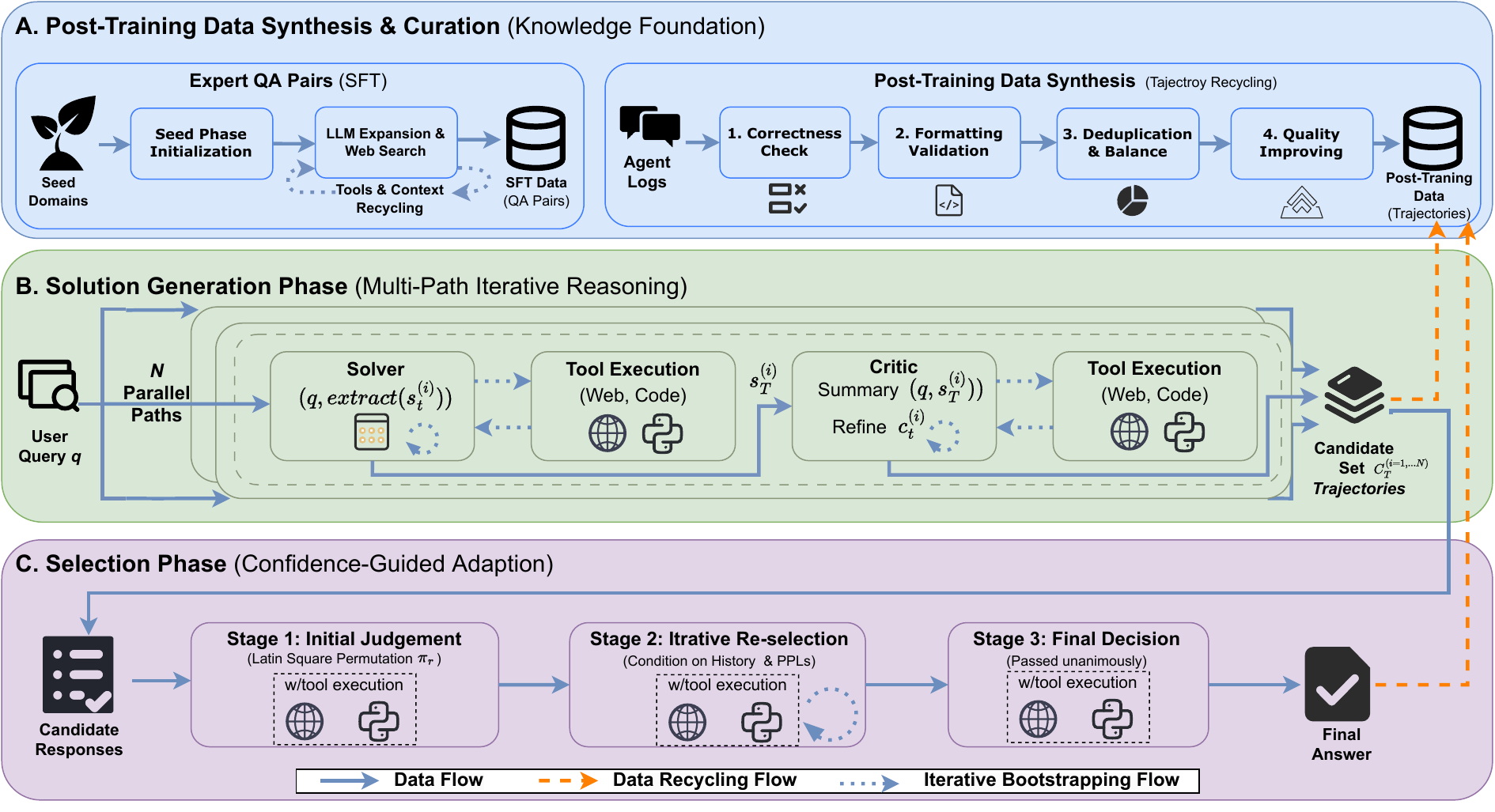}
\caption{\textbf{Overall Framework of ReThinker}: A Data-Driven and Uncertainty-Guided Agentic System for Expert-Level Scientific Reasoning. The framework comprises three integrated phases: \textbf{(A) Post-Training Data Synthesis \& Curation}, where trajectory recycling and a validation agent generate and refine expert QA pairs through correctness checks, formatting, deduplication, and quality balancing; \textbf{(B) Multi-Path Iterative Reasoning}, where parallel Solver-Critic paths execute tool-enhanced reasoning to produce candidate trajectories from user queries; and \textbf{(C) Confidence-Guided Selection}, a three-stage process employing Latin Square Permutation Test for initial judgment, iterative re-selection conditioned on historical data and perplexity scores (PPLs), and unanimous voting for final decision. The system features dual feedback loops—Data Recycling Flow and Iterative Bootstrapping Flow—that continuously enhance the knowledge foundation and reasoning capabilities.}
\label{fig:overall_workflow}
\end{figure*}

Figure~\ref{fig:overall_workflow} illustrates our data-driven, uncertainty-guided iterative reasoning framework.
The framework is organized into three tightly coupled phases, corresponding to the three panels.


\subsection{Post-Training Data Synthesis \& Curation}\label{sec:post_training}

Post-training data quality is as critical as agent workflow design for scientific reasoning.
Rather than relying on a single form of supervision, we decompose post-training data into two  complementary components: (1) expert-level QA pairs for supervised fine-tuning, and (2) adaptive trajectory utilization and recycling.


\subsubsection{Expert QA Pairs for Supervised Fine-Tuning}
To reduce human efforts and make the whole data synthesis pipeline more scalable and autonomous, as shown in Figure~\ref{fig:overall_workflow}A  (top-left), we propose the LLM-based multi-agent seed phrase initialization and online extraction to automatically construct seed phrases. They are then used to generate QA pairs, following the workflow proposed in WebExplorer~\citep{liu2025webexplorer}. Users only need to specify the interested topics such as biology and business, LLM agents will then propose initial seed noun phrases from them and extract and refine professional and uncommon noun phrases from trajectories and QA contexts during the subsequent flow. 

\textbf{Seed Domain Initialization.} Specifically, we utilize LLMs to generate 10 common phrases in 23 specific general domains from \texttt{natural sciences, social sciences, humanities, applied sciences}, and etc. As a result, we collect 230 very high-level and general seed phrases, including \texttt{natural selection, social stratification, color theory}, and \texttt{failure analysis}. These seed phrases are then used as the initial input for the whole automatic self-evolving agentic data synthesis pipeline to generate QA pairs and trajectories for subsequent model training.

\textbf{Automatic Seed Phrase Updating.} During the data synthesis stage proposed in ~\citep{liu2025webexplorer}, the searched web snippets and full contexts are only used once for QA generation and directly discarded after that. Due to the high cost of LLMs and web retrieval services, the aforementioned process leads to significant data generation expenses and inefficient utilization of retrieved content. 
Therefore, we recycle the previously discarded contexts during the data synthesis flow and extract seed phrases from them. The pool of seed phrases is online updated and extended using the searched web context, generated QA pairs, and trajectories.

\subsubsection{Adaptive Trajectory Utilization and Recycling}
\label{sec:adaptive_trajectory}
Reasoning trajectories constitute critical supervision signals and make a big difference to the performance of downstream agent.
Therefore,
as shown in Figure~\ref{fig:overall_workflow}A (top-right), we propose an adaptive trajectory synthesis and recycling framework that systematically governs how reasoning trajectories are generated, selected, and reused during post-training to make use of reasoning trajectories.
To this end, all collected trajectories (Agent Logs) are filtered, annotated, and curated offline using a combination of automatic metrics, such as answer correctness, tool-use efficiency, and reasoning coherence in support of enhancing the reasoning and tool-use capabilities of LLM-based agents in scientific research scenarios.

\textbf{Adaptive Trajectory Generation} Rather than relying on fixed-length reasoning chains or static tool-invocation policies, our framework enables adaptive control over reasoning trajectories. During question answering, the agent dynamically explores multiple candidate reasoning paths through iterative reasoning and tool interactions, selecting effective tool-call sequences and progressively refining intermediate hypotheses.

\textbf{Multi-Stage Data Quality Assurance Pipeline.} To ensure the reliability of generated trajectories, we design a multi-stage data quality assurance pipeline, which systematically transforms raw generated trajectories into a high-quality SFT dataset. 
\begin{itemize}[leftmargin=*]
\item \textbf{Correctness Check}. We first perform outcome-based filtering using a strong judge model to verify final answer correctness. Trajectories that fail to produce correct answers are discarded, preventing the model from inheriting erroneous reasoning or hallucinated solutions;

\item \textbf{Formatting Validation}. We enforce strict structural constraints on reasoning trajectories, such as: (i) \texttt{Answer Format}: the final result encapsulated within \verb|<answer></answer>| tag; (ii) \texttt{Interaction Integrity}: all dialogues must follow a consistent "User-Assistant" pairing and that no assistant response is empty; (iii) \texttt{Tool-Invocation Constraint}:To prevent inefficient reasoning, we filter trajectories based on tool-use density. We discard samples where the number of tool calls is either insufficient to resolve the query or excessively high but ineffective to solve problems.

\item \textbf{Deduplication \& Balance}. Redundant data are pruned to avoid overfitting on frequent patterns. Then, we rebalance data distribution across all reasoning phases, thereby mitigating model bias toward high-frequency patterns.

\item \textbf{Quality Improvement}. 
we finally assess the rationality and effectiveness of data. (i) \texttt{CoT-Response Alignment}: We filter out those data whose internal reasoning contradicts external outputs; (ii) \texttt{Successful Tool Execution}: Data which the model provides failed tool calling are excluded to ensure quality of the curated SFT data.

\end{itemize}



\subsection{Multi-Path Solution Generation}\label{sec:solution_generation}


As illustrated in Figure~\ref{fig:overall_workflow}B (middle), we instantiate $N \in \mathbb{Z}^{+}$ parallel reasoning paths, each consisting of solver and critic stages. Both solver and critic progressively improve solution quality through multi-round iterative rethinking. The critic stage is further equipped with guided reflection, which allows the critic to capture and correct subtle, fine-grained issues arising throughout the reasoning process. 



\textbf{Stage 1: Solver Stage with Rethinking.} Following prior work on iterative refinement~\citep{tian2025think,xu2025adaptive}, each path $i  (i = 1,...,N)$ performs $T_{solver}^{(i)} \in \mathbb{Z}^{+}$ rounds of reasoning. Each round invokes reasoning tools multiple times to retrieve relevant knowledge or verify reasoning steps, after which a single final answer is produced. This final answer is then extracted and fed into the subsequent round, prompting the model to reconsider its reasoning and iteratively refine the solution.

Let $s^{(i)}_{t}$ denotes the solution generated at round $t$, where $t = 0, \dots, T_i-1$. The solver stage can be represented as:
\begin{equation}
  s^{(i)}_{t+1} = \text{Solver}\bigl(q, \text{extract}(s^{(i)}_{t})),  
\end{equation}
where $q$ is the problem statement, $\text{extract}(s^{(i)}_{t}))$ denotes the final conclusion extracted from the reasoning trajectory $s^{(i)}_{t}$ of round $t$. This rethinking mechanism stably elevates reasoning quality, ensuring that easily correctable errors are eliminated before reflection.

\textbf{Stage 2: Critic Stage with Guided Reflection.}
Due to the context length limitations of LLMs, conventional reflection approaches over reasoning trajectories typically rely on partial outputs or or compressed representations, which may lead the reflection to overlook the fine-grained issues in the reasoning process. To address this limitation, we propose a guided reflection method. The reasoning trajectory produced by the Solver stage is first summarized into three components: summary of the trajectory, the final answer, and 
key areas for improvement. The Critic module then performs reflection based on these three components. Since the key areas for improvement are derived from the complete reasoning trajectory, this approach enables comprehensive analysis spanning fine-grained issues as well as high-level logical flaws.

The summary process can be represented as:
\begin{equation}
     y^{(i)}, a^{(i)}, k^{(i)} = \text{Summary}(q, s_{T_i}^{(i)}),
\end{equation}
where $y^{(i)}, a^{(i)}$ and $k^{(i)}$ denote the key reasoning steps, the final answer, and the key areas for improvement extracted from the Solver’s reasoning trajectory of the last round $s_{T_i}^{(i)}$ for path $i$, respectively. Let $c^{(i)}_{t}$ denotes the critic result at round $t$, where $c^{(i)}_{t}), t=0, \dots,T_{critic}^{(i)}-1$. The critic stage can be represented as:
\begin{equation}
    c^{(i)}_{t + 1} = \text{Critic}\left( q, y^{(i)}, a^{(i)}, k^{(i)},  \text{extract}(c^{(i)}_{t}) \right) .
\end{equation}



\subsection{Confidence-Guided Selection}
\label{sec:selection}
As shown in Figure~\ref{fig:overall_workflow}C (bottom), we adopt a three-stage \emph{confidence-guided evaluation} framework.
The selector first scores all candidates with confidence estimates, then iteratively refines its selection using perplexity-weighted confidence under Latin-square permutations to eliminate position bias.
A final aggregation step is applied only when cross-round selections are inconsistent.
This design concentrates computation on uncertain cases while remaining robust to ordering effects and early-round noise.

The solution generation stage produces a candidate set $\mathcal{C} = \{c_1, c_2, ..., c_n\}$ of feasible answers, each accompanied by a reasoning trajectory. The \textbf{selector} must identify the optimal answer while mitigating systematic errors from single-pass inference and position bias. We frame this as a three-stage confidence-calibrated decision process.

\textbf{Stage 1: Initial Judgement.}
The problem statement $q$ and candidate set $\mathcal{C}$ are formatted into a structured prompt that elicits both a preliminary selection $s_0 \in \mathcal{C}$ and a confidence estimate. Crucially, to eliminate ordering bias, we permute candidate positions via Latin squares: for round $r$, we apply a permutation $\pi_r$ drawn from a pre-computed Latin square $\mathcal{L}$, presenting candidates as $(\pi_r(c_1), \pi_r(c_2), ..., \pi_r(c_n))$. This ensures each candidate appears equally often in every position across rounds, forcing the model to focus on content rather than ordinal heuristics.

\textbf{Stage 2: Iterative Re-selection.} 
The initial judgement's \textbf{perplexity} $\text{PPL}(s_0)$ serves as a gating signal for progressive refinement. Perplexity is computed as:
\begin{equation}
    \text{PPL}(s_0) = \exp\left(-\frac{1}{T_{seq}}\sum_{t=1}^{T_{seq}} \log p_{\theta}(x_t \mid x_{<t})\right),
\end{equation}
where $x_t$ are tokens in the selection rationale and $T_{seq}$ is the sequence length. High PPL indicates uncertainty, triggering $R$ additional re-selection rounds. In each round $r$, the model conditions on the \textit{aggregated history} $H_r = \{s_0, s_1, ..., s_{r-1}\}$ and their PPL scores, enabling \textbf{confidence-weighted progressive refinement} where selections become increasingly precise. The process amplifies high-certainty choices while suppressing noisy candidates through Bayesian updating of selection probabilities.

\textbf{Stage 3: Final Decision.} 
We synthesize historical selections to produce a definitive answer. Let $\mathcal{C}_{\text{hist}} = \{c \in \mathcal{C}: \exists\, r \in \{0,...,R\} \text{ s.t. } s_r = c\}$ be the set of candidates ever selected. If $|\mathcal{C}_{\text{hist}}| = 1$, the answer is output directly, bypassing this stage. Otherwise, we execute a \textbf{final adjudication pass} that conditions on 
this candidate set with responded answers and confident scores, discarding never-mentioned candidates and resolving inconsistencies through a decisive selection. This ensures robust aggregation with each historically-selected candidate treated as an independent option.

\section{Experiments}\label{sec:experiments}

\subsection{Experiment Setup}

We evaluate our method on three representative and challenging reasoning benchmarks, which comprehensively assess advanced analytical and agentic reasoning capabilities:

\begin{itemize}
    \item \textbf{Humanity's Last Exam (HLE)}~\cite{phan2025humanity}:
    A large-scale expert-level benchmark with challenging problems across diverse scientific fields. It tests whether AI systems can demonstrate deep reasoning and knowledge at near-human expert levels. Following prior work, We evaluate on a text-only subset of 2158 validation instances~\cite{team2025mirothinker}.

    \item \textbf{GAIA}~\cite{mialon2023gaia}: 
    A benchmark composed of real-world tasks that require tool usage, web navigation, and multi-step planning. Following prior work, we evaluate on a text-only subset of 103 validation instances~\cite{li2025webthinker,wu2025webdancer}.
    
    \item \textbf{XBench-DeepSearch}~\cite{chen2025xbench}: 
    A professionally-aligned benchmark that focuses on evaluating AI agent's tool usage capabilities, specifically in deep information retrieval and complex search tasks. And it totally contains 100 expert-level reasoning problems.
\end{itemize}


\paragraph{Evaluation Protocol.}
All benchmarks are evaluated using an LLM-as-a-Judge framework. Specifically, GAIA and XBench-DeepSearch are evaluated using \textit{gpt-4.1-2025-04-14}, while HLE follows its official evaluation protocol with judgments produced by \textit{o3-mini-2025-01-31}.


\begin{table}[htbp]
  \centering
  \caption{Main Results of Inference Accuracy (\%) on Expert-Level Reasoning Benchmarks.}
  \renewcommand{\arraystretch}{1.15}
  \begingroup 
  \let\oldcite\cite 
  \renewcommand{\cite}[1]{{\scriptsize\oldcite{#1}}} 
  \begin{tabularx}{0.5\textwidth}{Xccc}
    \toprule
    Benchmarks & HLE & GAIA & XBench \\
    \midrule
    \multicolumn{4}{l}{\textbf{Foundation Models with Tools}} \\
    \midrule
    Kimi K2~\cite{team2025kimi} & 18.1 & 57.7 & 50.0 \\
    Claude-4.5-Sonnet~\cite{Anthropic2025Claude} & 24.5 & 71.2 & 66.0 \\
    DeepSeek-V3.2~\cite{liu2025deepseek} & 27.2 & 63.5 & 71.0 \\
    GLM-4.6~\cite{Zhipu2025GLM-4.6} & 30.4 & 71.9 & 70.0 \\
    GPT-5-high~\cite{OpenAI2025GPT5} & 35.2 & 76.4 & 77.8 \\
    Gemini-3-Pro~\cite{google2025gemini3} & 38.3 & 79.0 & 87.0 \\
    \midrule
    \multicolumn{4}{l}{\textbf{Inference Frameworks}} \\
    \midrule
    WebExplorer~\cite{liu2025webexplorer} & 17.3 & 50.0 & 53.7 \\
    OpenAI DeepResearch~\cite{OpenAI2025DeepResearch} & 26.6 & 67.4 & -- \\
    Kimi Researcher~\cite{Kimi2025DeepResearch} & 26.9 & -- & 69.0 \\
    Tongyi DeepResearch~\footnotesize{(30B-A3B)}~\cite{team2025tongyi} & 32.9 & 70.9 & 75.0 \\
    MiroThinker-v1.0~\footnotesize{(30B)}~\cite{team2025mirothinker} & 33.4 & 73.5 & 70.6 \\
    \midrule
     ReThinker~\footnotesize{(OpenPangu-72B)} & 33.1 & 72.8 & 78.0 \\
    ReThinker~\footnotesize{(Gemini-3-Pro)} & \textbf{52.2} & \textbf{81.6} & \textbf{90.0} \\
    \bottomrule
  \end{tabularx}
  \endgroup 
  \label{tab:main_results}
  \vspace{-0.4cm}
\end{table}

\subsection{Main Results}

Table~\ref{tab:main_results} summarizes the main experimental results on three text-only reasoning benchmarks.
Overall, our ReThinker framework consistently outperforms both foundation models with tools and existing inference frameworks across all benchmarks.

On \textbf{HLE}, ReThinker instantiated with Gemini-3-Pro achieves \textbf{52.18\%} accuracy, substantially surpassing all baselines.
Compared to strong tool-augmented foundation models such as OpenAI-GPT-5-high (35.2\%) and Gemini-3-Pro used directly (38.3\%), our approach yields improvements of \textbf{16.9} and \textbf{13.8} percentage points, respectively.
It also significantly outperforms specialized inference frameworks, including Tongyi DeepResearch (32.9\%) and MiroThinker-v1.0 (33.4\%), demonstrating the effectiveness of adaptive trajectory utilization and confidence-guided selection for high-difficulty scientific reasoning.

On \textbf{GAIA}, ReThinker (Gemini-3-Pro) achieves \textbf{81.55\%} accuracy, establishing a new state of the art among all compared methods.
This result exceeds Gemini-3-Pro with tools (79.0\%) and other deep research systems such as Tongyi DeepResearch (70.9\%) and MiroThinker-v1.0 (73.5\%), validating the robustness of our framework in complex, tool-intensive, real-world tasks.

On \textbf{XBench-DeepSearch}, our method reaches \textbf{90.0\%} accuracy, outperforming all open-source baselines and improving upon Gemini-3-Pro with tools (87\%).
These gains indicate that ReThinker not only enhances answer correctness but also provides more stable and reliable reasoning under expert-level evaluation settings.

Taken together, the results demonstrate that our framework consistently amplifies the reasoning capabilities of strong foundation models, particularly on benchmarks that demand long-horizon planning, multi-step inference, and precise tool orchestration, rather than shallow retrieval or memorization.

\subsection{Component Analysis}

To quantify the contribution of each component in our framework, we conduct a controlled component analysis on a representative subset of \textbf{500 text-only HLE problems}, sampled from the full benchmark with matched category distribution.
We adopt a modular decoupling strategy to isolate the effect of each stage. To ensure fair comparison, all variants are instantiated with \textbf{OpenPangu} as a unified backbone.

\textbf{Solver Phase: Re-Answer Synthesis Improves Initial Solution Quality.}
As shown in Table~\ref{tab:solver_gain}, introducing multi-round re-answer synthesis yields a \textbf{1.4\%} absolute improvement in Pass@5. This gain is achieved by iteratively bootstrapping candidate solutions across rounds, allowing the solver to refine earlier reasoning traces. Although the numerical improvement is modest, it plays a critical role in reducing low-level errors and narrowing the error surface exposed to downstream modules. As a result, the subsequent \textit{Critic} phase can focus on high-order logical inconsistencies rather than correcting superficial or syntactic mistakes. 

\begin{table}[htbp]
  \centering
  \caption{Effect of Re-Answer Synthesis in the Solver Phase.}
    \begin{tabular}{rlc}
    \toprule
    \multicolumn{1}{l}{Phase} & Stage & \textcolor[rgb]{ 0,  0,  1}{Pass@5} \\
    \midrule
    \multicolumn{1}{r}{\multirow{2}[2]{*}{Solver}} & Initial Solver & 38.00\% \\
          & Re-Answer Solver & 39.40\% \\
    \bottomrule
    \end{tabular}%
  \label{tab:solver_gain}%
\end{table}%

\textbf{Critic Phase: Guided Reflection with Structured Summary Is the Primary Contributor.}
Table~\ref{tab:critic_gain} demonstrates that the Critic phase delivers the most substantial single-stage improvement, contributing a \textbf{3.8\%} Pass@5 gain over the solver output.
Notably, \emph{Critic with Summary \& Guidance} outperforms both the \emph{Final Answer-only} and \emph{Summary-only} variants by \textbf{2.8\%} and \textbf{1.6\%}, respectively.
This result confirms that structured guidance is essential for effective reflection. 


\begin{table}[htbp]
  \centering
  \caption{Impact of Guided Reflection Strategies in the Critic Phase.}
    \begin{tabular}{rlc}
    \toprule
    \multicolumn{1}{l}{Phase} & Setting & \textcolor[rgb]{ 0,  0,  1}{Pass@5} \\
    \midrule
    \multicolumn{1}{l}{Solver} & Re-Answer Solver & 39.40\% \\
    \midrule
    \multicolumn{1}{r}{\multirow{3}[2]{*}{Critic}} & Critic w/Final Answer & 40.40\% \\
          & Critic w/Summary & 42.00 \% \\
          & Critic w/Summary \& Guidance & 43.20\% \\
    \bottomrule
    \end{tabular}%
  \label{tab:critic_gain}%
\end{table}%

\textbf{Selector Phase: Compound Gains from Confidence Guidance and Position Robustness.}
As reported in Table~\ref{tab:select_gain}, the Selector phase produces a cumulative \textbf{5.6\%} improvement in hit rate and a corresponding \textbf{5.6\%} gain in Pass@1 through progressive refinement.
The stage-wise improvements reveal complementary effects:
\begin{itemize}
    \item \textbf{Initial Judgement} establishes a \textbf{65.27\%} hit rate baseline, comparable to naive best-of-$N$ selection.
    
    \item \textbf{+Iterative Judgement} improves hit rate by \textbf{2.78\%}, indicating that re-conditioning on prior selections effectively filters spurious candidates even without explicit confidence modeling.
    
    \item \textbf{+Perplexity Guidance} yields an additional \textbf{1.39\%} gain, validating perplexity as a reliable uncertainty signal. This mechanism constitutes the core of our test-time scaling strategy, allocating additional compute to instances where the model exhibits higher uncertainty.
    
    \item \textbf{+Latin Square Rank} contributes the final \textbf{1.39\%} Pass@1 improvement, demonstrating that position bias is non-negligible in selection. By enforcing uniform rank exposure across rounds, this strategy ensures that selection decisions are driven by content quality rather than ordinal position.
\end{itemize}

\begin{table}[htbp]
  \setlength{\tabcolsep}{4pt} 
  \centering
  \caption{Incremental Gains from Confidence-Guided Selection.}
    \begin{tabular}{rlrc}
    \toprule
    \multicolumn{1}{l}{Phase} & Setting & \multicolumn{1}{l}{Hit Rate} & \textcolor[rgb]{ 0,  0,  1}{Pass@1} \\
    \midrule
    \multicolumn{1}{r}{\multirow{4}[2]{*}{Selector}} & Initial Judgement & 65.27\% & 28.20\% \\
          & +Iterative Judgement & 68.05\% & 29.40\% \\
          & +Perplexity Guidance & 69.44\% & 30.00\% \\
          & +Latin Square Rank & 70.83\% & 30.60\% \\
    \bottomrule
    \end{tabular}%
  \label{tab:select_gain}%
\end{table}%

\section{Discussion and Analysis}

Building upon the effectiveness results in Section~\ref{sec:experiments}, we further examine ReThinker’s efficiency and operational characteristics from three complementary perspectives:
(1) \textbf{Tool Use Statistics}, which quantify the average number of tool invocations per problem across phases (Solver, Critic, and Selector);
(2) \textbf{Solver-to-Critic Benefits}, which analyze how phased refinement improves solution quality and task adaptation;
and (3) \textbf{The Guidance of Perplexity}, which evaluates the statistical and behavioral impact of perplexity-guided decision making in the Selector.

\subsection{Tool Use Statistics}

\begin{figure}[htbp]
\begin{center}
\centerline{\includegraphics[width=0.45\textwidth]{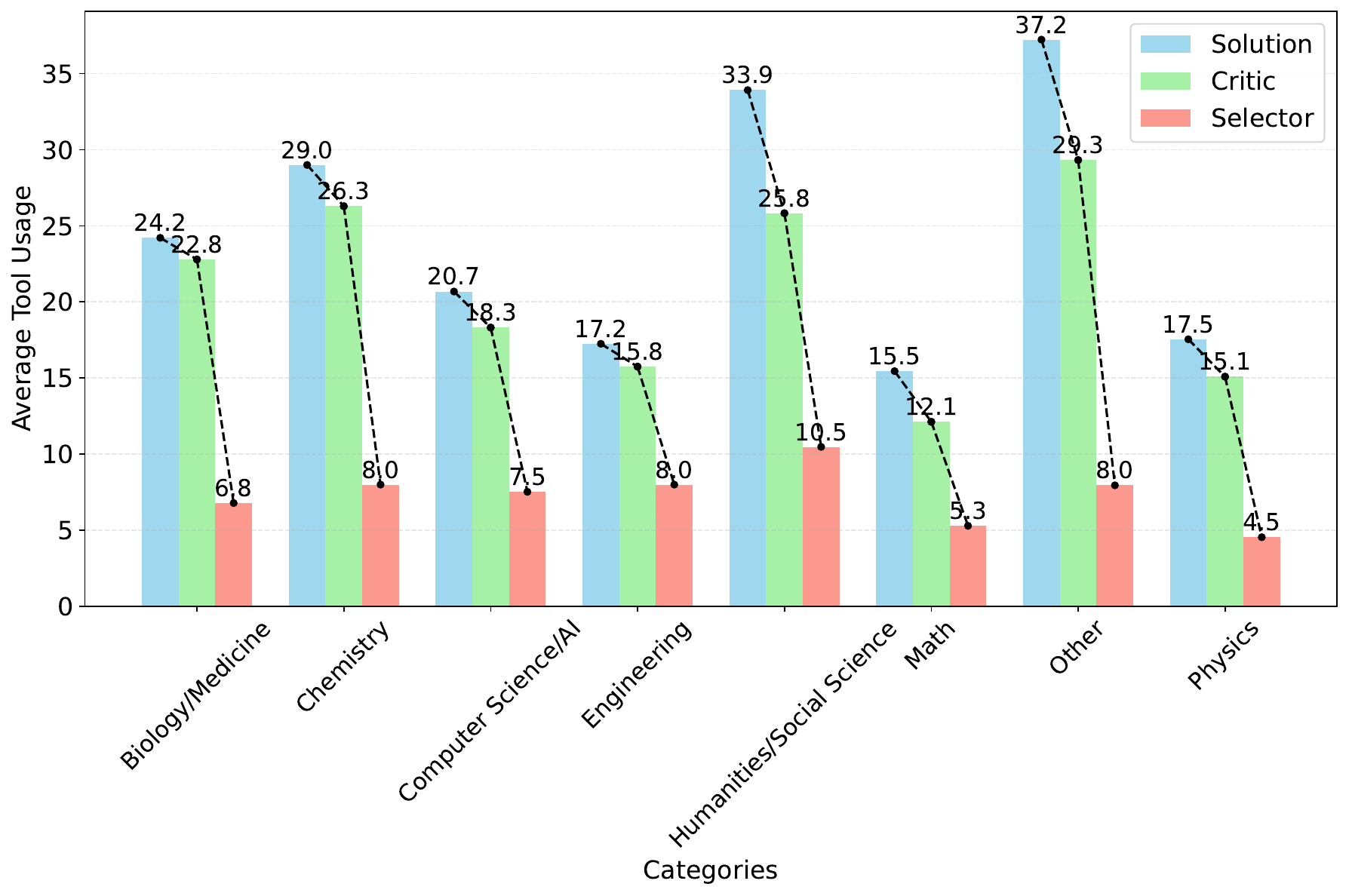}}
\caption{Tool Usage Statistics across Reasoning Phases in ReThinker.}
\label{fig:tool_usage}
\end{center}
\vspace{-0.5cm}
\end{figure}

Figure~\ref{fig:tool_usage} shows a clear and monotonic decrease in tool invocation from the Solver to the Critic and finally to the Selector phase.
The \textbf{Solver} phase exhibits the highest tool usage, reflecting its role as the primary exploration and information acquisition stage.
At this stage, the model operates under maximal uncertainty and actively queries external tools to construct an initial knowledge foundation.

Upon transitioning to the \textbf{Critic} phase, average tool usage decreases by a factor of \textbf{3.72}, indicating that the structured summary and critique mechanism effectively consolidates context and localizes residual knowledge gaps, rather than re-exploring the problem space broadly.
By the \textbf{Selector} phase, tool calls drop to single-digit levels, and final decisions are made almost entirely based on internal confidence signals.

This monotonic decline demonstrates that ReThinker successfully accumulates, compresses, and reuses external information across its reasoning trajectory.
The observed trend validates our design hypothesis: early-stage exploration is resource-intensive but necessary, while later-stage refinement and selection increasingly rely on synthesized internal representations, thereby minimizing external dependencies while improving decision confidence.

\subsection{Solver-to-Critic Benefits}
\begin{figure}[htbp]
\begin{center}
\centerline{\includegraphics[width=0.45\textwidth]{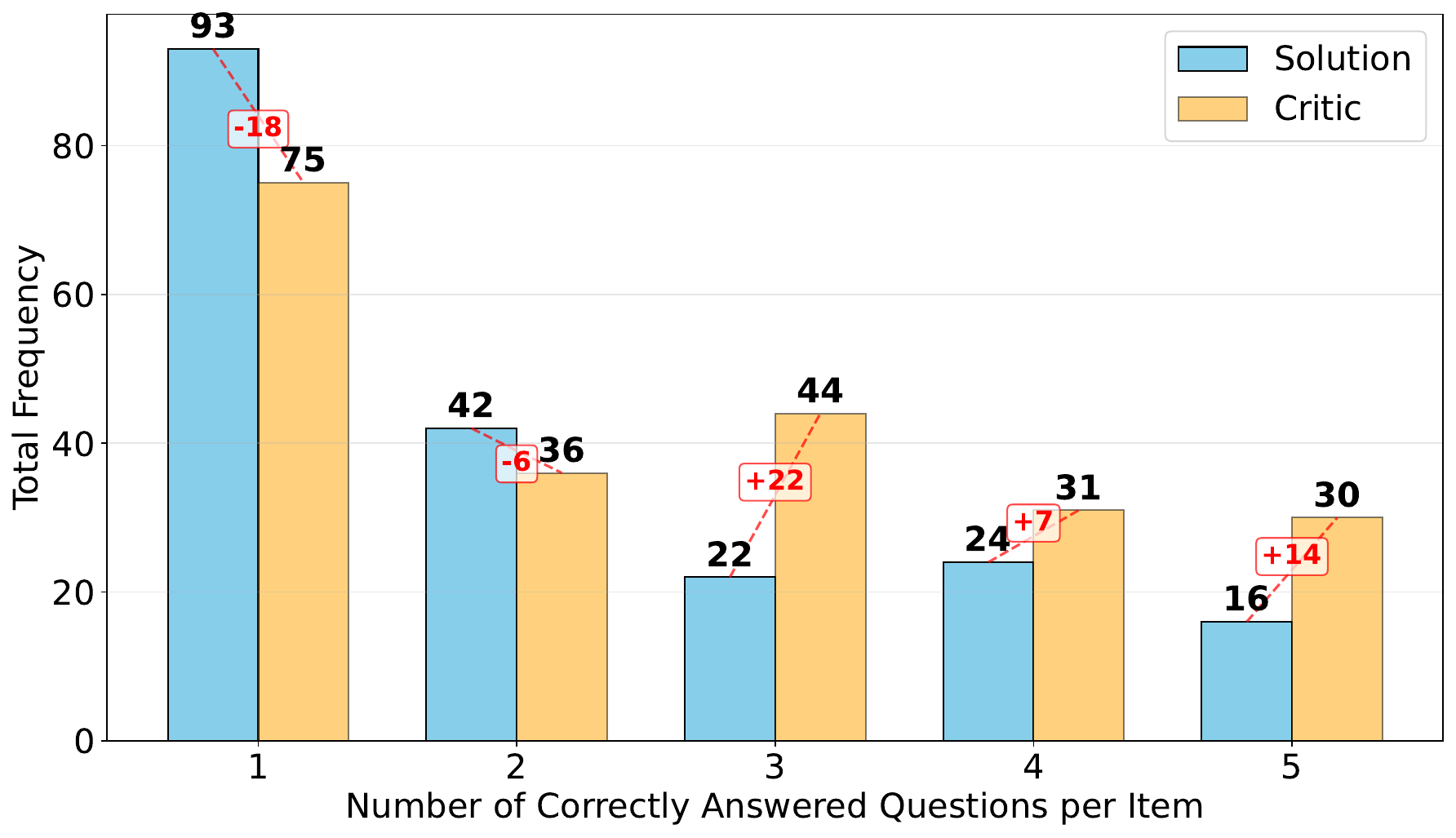}}
\caption{Distributional Shift in Correct-Answer Trajectories from Solver to Critic.}
\label{fig:solver_to_critic_benefits}
\end{center}
\vspace{-0.5cm}
\end{figure}

Figure~\ref{fig:solver_to_critic_benefits} illustrates the distributional shift in correct-answer trajectories between the Solver and Critic phases. 
In the \textbf{Solver} phase, the distribution is highly skewed: 93 problems yield only 1 correct candidate out of 5 generated paths, while only 16 problems achieve the ideal 5/5 correct rate.
This reflects the Solver’s role as an exploratory generator, producing diverse but noisy hypotheses with limited self-correction.

After transitioning to the \textbf{Critic} phase, the distribution shifts toward higher-quality regions.
The number of problems with only a single correct answer decreases from 93 to 75, while those achieving 5 correct answers nearly double to 30.
Correspondingly, the mean number of correct answers increases from \textbf{2.1} (Solver) to \textbf{2.6} (Critic).

These results indicate that the Critic does not merely filter existing candidates, but actively recalibrates the solution ensemble.
Guided reflection systematically uplifts marginal trajectories, converting previously weak or partially correct solutions into viable answers.
This ensemble-level redistribution highlights the Critic’s role as a global quality amplifier rather than a local verifier.

\subsection{The Guidance of Perplexity}

\begin{figure}[htbp]
\begin{center}
\centerline{\includegraphics[width=0.45\textwidth]{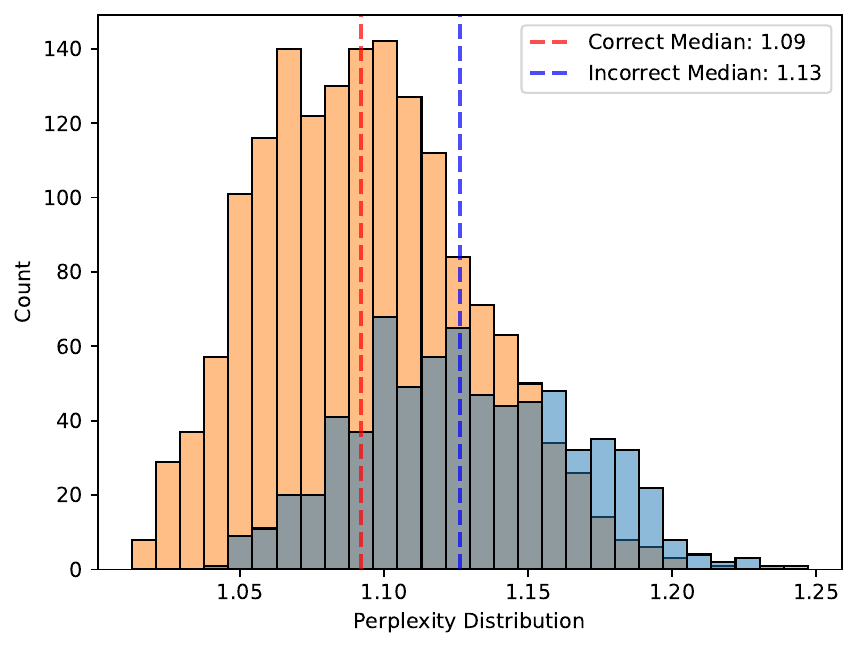}}
\caption{Separation between Correct and Incorrect Answers Induced by Perplexity.}
\label{fig:guidance_ppl}
\end{center}
\vspace{-0.5cm}
\end{figure}

Figure~\ref{fig:guidance_ppl} visualizes the empirical relationship between perplexity and answer correctness across four selector iterations.
Correct answers predominantly cluster at lower perplexity values, while incorrect answers exhibit a pronounced rightward shift, forming a clear separation between high- and low-confidence regions.
This monotonic pattern confirms that perplexity serves as a reliable proxy for model uncertainty during selection.

Table~\ref{tab:selector_iter} further quantifies the effect of perplexity-guided re-selection across iterative rounds, measured by the cumulative number of correctly selected answers.
Starting from an identical initial baseline of 141 correct selections, the two settings (with and without perplexity guidance) diverge immediately.
In Round 1, the perplexity-guided selector gains \textbf{7} correct selections, whereas the non-guided variant incurs a net loss, indicating that confidence-agnostic re-selection amplifies noise rather than signal in early iterations.

\begin{table}[htbp]
  \setlength{\tabcolsep}{2pt} 
  \centering
  \caption{Effect of Perplexity-Guided Re-Selection across Iterative Selector Rounds.}
    \begin{tabular}{cccccc}
    \toprule
    \multirow{2}{*}{Selector} & \multicolumn{1}{c}{\multirow{2}{*}{\makecell[c]{Initial \\ Judgement}}} & \multicolumn{4}{c}{Number of Iteration} \\
    \cmidrule{3-6}
    & & 1 & 2 & 3 & 4 \\
    \midrule
    w/PPL & \multirow{2}{*}{141} & 148 (↑) & 150 (-) & 150 (-) & 153 (↑) \\
    wo/PPL & & 140 (↓) & 144 (↑) & 143 (↓) & 147 (↑) \\
    \bottomrule
    \end{tabular}%
  \label{tab:selector_iter}%
\end{table}%

The temporal dynamics reveal distinct convergence behaviors.
The perplexity-guided selector exhibits steady improvement followed by clear saturation, while the non-guided variant displays volatile oscillations reminiscent of a random walk.
These results validate our core hypothesis: perplexity is not merely a diagnostic metric, but an actionable control signal that allocates the selector’s computational budget—intensifying refinement where uncertainty remains high and terminating early when confidence is sufficient.

\section{Conclusion}

In this paper, we propose ReThinker, an uncertainty-gated orchestration framework for scientific reasoning tasks, and design a stage-wise solver-critic-selector architecture. ReThinker can learn from synthesized reasoning trajectories and significantly improve inference efficiency and accuracy. ReThinker can also acquire strong zero-shot transfer ability across expert-level benchmarks and yield an effective initialization for few-shot adaptation to unseen scientific domains.



\bibliography{example_paper}
\bibliographystyle{icml2025}

\newpage
\appendix
\onecolumn

\section{Appendix}

\subsection{Scientific Reasoning Benchmarks for LLMs}

Recent benchmarks have been proposed to evaluate the reasoning capabilities of large language models across expert-level scientific knowledge, open-world problem solving, and executable tool use. Below we detail three representative benchmarks with quantifiable distributions.

\subsubsection{Humanity’s Last Exam~(HLE)}

\textbf{Humanity's Last Exam (HLE)}~\cite{phan2025humanity} is an expert-level scientific benchmark explicitly designed to resist shallow retrieval and pattern matching. It comprises 2,158 text-only validation questions spanning over 100 academic disciplines, requiring deep domain expertise, multi-step causal reasoning, and precise logical inference. Unlike traditional knowledge benchmarks where frontier models exceed 90\% accuracy, HLE presents a significant challenge with most models scoring below 10\%.

The dataset emphasizes \textit{anti-retrieval} characteristics through two primary question formats: 24\% multiple-choice questions requiring nuanced discrimination among highly plausible distractors, and 76\% exact-match short-answer questions demanding precise symbolic or conceptual responses. Table~\ref{tab:hle_display} presents the domain distribution, with Mathematics comprising the largest proportion (45.23\%, 976 questions), followed by Computer Science/AI (10.38\%) and Biology/Medicine (10.29\%). Notably, the benchmark exhibits a substantial performance gap between human experts (average accuracy $>$90\%) and state-of-the-art models (Grok-4 achieves $\sim$25.4\%, while GPT-4 and Claude-3 score $<$10\%).

\begin{table}[htbp]
  \vspace{-1em}
  \centering
  \caption{HLE Dataset Composition and Distribution (text-only)}
    \begin{tabular}{lrr}
    \toprule
    Category & \multicolumn{1}{l}{Number of Data} & \multicolumn{1}{l}{Proportion (\%)} \\
    \midrule
    Biology/Medicine & 222   & 10.29 \\
    Chemistry & 101   & 4.68 \\
    Computer Science/AI & 224   & 10.38 \\
    Engineering & 64    & 2.97 \\
    Humanities/Social Science & 193   & 8.94 \\
    Math  & 976   & 45.23 \\
    Other & 176   & 8.16 \\
    Physics & 202   & 9.36 \\
    \midrule
    Total & 2158  & 100.00 \\
    \bottomrule
    \end{tabular}%
  \label{tab:hle_display}%
  \vspace{-1em}
\end{table}%


\subsubsection{GAIA}
\textbf{GAIA}~\cite{mialon2023gaia} focuses on \textit{open-world reasoning} and \textit{tool-assisted problem solving} through 103 text-based validation tasks specifically curated to require multi-step planning, information synthesis, and interaction with external tools such as web browsers and calculators. The benchmark emphasizes \textit{grounded reasoning} under realistic constraints, systematically exposing limitations in long-horizon planning and reliable tool orchestration.

\begin{table}[htbp]
  \centering
  \caption{Introduction of GAIA 103 Validation (text-only)}
    \begin{tabular}{lcccc}
    \toprule
    Difficulty & Question Count & Proportion (\%) & Avg. Human Steps & Primary Tool Requirements \\
    \midrule
    Level 1 & 39    & 37.86 & \texttt{<} 5 steps & Minimal \\
    Level 2 & 52    & 50.49 & 5-10 steps & Web Search+Calculator \\
    Level 3 & 12    & 11.65 & \texttt{>} 10 steps & Multi-Tool Orchestration \\
    \midrule
    Total & 103   & 100   & -     & - \\
    \bottomrule
    \end{tabular}%
  \label{tab:gaia_display}%
\end{table}%

GAIA stratifies tasks into three difficulty tiers based on the complexity of required reasoning chains and tool dependencies (Table~\ref{tab:gaia_display}). Level 1 (37.86\%, 39 tasks) requires $<$5 reasoning steps with minimal tool usage; Level 2 (50.49\%, 52 tasks) demands 5--10 steps incorporating web search and calculation; Level 3 (11.65\%, 12 tasks) necessitates $>$10 steps with complex multi-tool orchestration.

\subsubsection{XBench-DeepSearch}

\textbf{XBench-DeepSearch} (Chinese version) is a professionally curated benchmark designed to evaluate the deep search capability of AI agents in real-world, open-domain environments. Each question requires multi-step information retrieval, cross-source reasoning, and synthesis, rather than direct fact lookup. The dataset is constructed and continuously refreshed by domain experts under an evergreen evaluation protocol, ensuring long-term validity and resistance to data contamination.

A standard release of XBench-DeepSearch consists of 100 questions, with problem types distributed to balance search breadth, reasoning depth, and practical task realism. Questions are intentionally heterogeneous, spanning multiple cognitive and operational demands commonly encountered by real-world AI agents.

\begin{table}[htbp]
  \centering
  \caption{Introduction of Xbench-DeepSearch (text-only)}  
    \begin{tabular}{lcp{8cm}}  
    \toprule
    Topic Domain & Number of Tasks & Typical Examples \\
    \midrule
    Business \& Finance & 12    & Stock exchanges (Shanghai Gold, Shenzhen), Economic indicators (GDP per capita), Corporate history (Alibaba founders), Brand analysis (Arc'teryx, Balenciaga), Market transactions \\
    \addlinespace
    Current Affairs \& Politics & 6     & International relations (Artemis Accords, defense agreements), Olympic medal adjustments, Border geography (Northeast China), Political history (Singapore founding), Military history (Nimitz-class carriers) \\
    \addlinespace
    Education \& Academia & 9     & Academic institutions (HKU faculty, U of T programs), Educational systems (Central Conservatory grading), Academic publications (CVPR papers, Nobel laureates), Historical academic comparisons \\
    \addlinespace
    Entertainment \& Media & 31    & Variety shows (``Farewell My Love 4'', ``Comedy Night''), Music (Grammy Awards, Taylor Swift analysis), Gaming (Black Myth: Wukong, Arknights), Film analysis (Oscar winners, Studio Ghibli), Bilibili content \\
    \addlinespace
    Geography \& Transportation & 13    & Beijing/Shanghai/Suzhou subway systems, Aviation (Beijing to Sydney flights), Urban landmarks (Three-monastery equidistant point), Railway schedules, Geographic information systems \\
    \addlinespace
    Humanities \& Social Sciences & 11    & Classical literature (Strange Stories from a Chinese Studio, Jin Yong novels), Historical artifacts (Tang Dynasty contracts), Cultural heritage (Porcelain Palace), Cuisine history, Historical events \\
    \addlinespace
    Natural Sciences & 3     & Physical chemistry (metal melting points, Tyndall effect), Traditional Chinese medicine (Compendium of Materia Medica) \\
    \addlinespace
    Sports & 7     & Competitive gaming (Dota2 TI, Esports), Professional sports (NBA, UEFA Champions League), Board games (Go, Snooker), Olympic swimming records \\
    \addlinespace
    Technology \& Engineering & 8     & Computer science (Java API, GPU FLOPS), Artificial intelligence (DeepSeek, OpenAI Codex), Autonomous driving (Didi/Volvo specs), UAV technology (DJI drones), Hardware specifications \\
    \midrule
    Total & 100   & -- \\
    \bottomrule
    \end{tabular}%
  \label{tab:xbench_display}%
\end{table}%


\subsection{Tool Details}
\label{sec:tools}

LLM agents in each phase is equipped with a Python interpreter,  pre-configured with three specialized tools: \texttt{web\_search}, \texttt{web\_parse} and \texttt{execute\_python\_code}. Table~\ref{tab:tool_details} provides a detailed description of each tool.

\begin{table}[h!]
\centering
\caption{Detailed descriptions of the tools available to LLM agents.}
\label{tab:tool_details}
\begin{tabular}{@{} l l p{8cm} @{}}
\toprule
\textbf{Tool} & \textbf{Syntax} & \textbf{Description} \\
\midrule

\texttt{web\_search} & \texttt{web\_search(keywords)} & 
Leverages the SERPER API to perform a web search based on the provided \texttt{keywords}. It returns a list of relevant URLs and their corresponding snippets. \\
\addlinespace

\texttt{web\_parse} & \texttt{web\_parse(link, query)} & 
Extracts targeted information from a webpage. First, it employs the JINA API to parse the content of the given \texttt{link} into Markdown format. Subsequently, an LLM is invoked to extract and synthesize the content most relevant to the \texttt{query} from the parsed text. \\
\addlinespace

\texttt{exec\_code} & \texttt{execute\_python\_code(code, timeout)} & 
Executes Python code asynchronously within a thread pool executor with configurable timeout (defaulting to 3600 seconds). It captures execution output, error messages, and runtime duration. When tracing is enabled in configuration, it incrementally persists execution records---including contextual metadata (query ID, payload), source code, output, and error streams---to a JSONL file using asynchronous I/O with file locking for thread-safe audit trails. \\

\bottomrule
\end{tabular}
\end{table}

\section{Experiments Details}

\subsection{Additional Experiments}

\subsubsection{Performance on Humanity's Last Exam}

\textbf{Overall Performance.} As shown in Table~\ref{tab:hle_performance}, the ReThinker framework demonstrates substantial improvements across all categories when powered by Gemini-3-Pro compared to OpenPangu-72B. On aggregate metrics, Gemini-3-Pro achieves a Pass@5 of \textbf{61.49\%} and Pass@1 of \textbf{52.18\%}, significantly outperforming OpenPangu-72B's \textbf{43.42\%} and \textbf{33.09\%}, respectively. The Hit Rate—measuring the proportion of problems where at least one solution is correct—increases from 76.20\% to 84.85\%, indicating superior solution coverage with the stronger base model.

\newcommand{\best}[1]{\textbf{#1}}

\begin{table}[htbp]
  \centering
  \caption{Performance comparison of ReThinker with different base models on HLE benchmark across categories.}
  \label{tab:hle_performance}
  \small
  \begin{tabular}{
    @{} 
    l 
    c c c 
    c c c 
    @{}}
    \toprule
    \multirow{2}{*}{Category} & 
    \multicolumn{3}{c}{ReThinker (OpenPangu-72B)} & 
    \multicolumn{3}{c}{ReThinker (Gemini-3-Pro)} \\
    \cmidrule(lr){2-4} \cmidrule(lr){5-7}
    & {Pass@5 (\%)} & {Pass@1 (\%)} & {Hit Rate (\%)} & {Pass@5 (\%)} & {Pass@1 (\%)} & {Hit Rate (\%)} \\
    \midrule
    Biology/Medicine & 39.64 & 29.73 & 75.00 & \best{55.86} & \best{43.69} & \best{78.23} \\
    Chemistry & 38.61 & 25.74 & 66.67 & \best{55.45} & \best{45.54} & \best{82.14} \\
    Computer Science/AI & 32.59 & 25.45 & \best{78.08} & \best{56.25} & \best{42.86} & 76.19 \\
    Engineering & 18.75 & 12.50 & 66.67 & \best{42.19} & \best{39.06} & \best{92.59} \\
    Humanities/Social Sci. & 51.30 & 43.01 & 83.84 & \best{67.88} & \best{58.55} & \best{86.26} \\
    Math & 47.85 & 37.81 & 79.01 & \best{65.16} & \best{58.30} & \best{89.47} \\
    Other & 51.70 & 38.64 & 74.73 & \best{70.45} & \best{57.39} & \best{81.45} \\
    Physics & 33.66 & 18.32 & 54.41 & \best{50.99} & \best{39.11} & \best{76.70} \\
    \midrule
    Average & 43.42 & 33.09 & 76.20 & \best{61.49} & \best{52.18} & \best{84.85} \\
    \bottomrule
  \end{tabular}
\end{table}

\textbf{Category-Specific Analysis.} The performance gap is particularly pronounced in Engineering, where Gemini-3-Pro achieves a \textbf{92.59\%} Hit Rate versus \textbf{66.67\%} for OpenPangu-72B, alongside a dramatic improvement in Pass@5 \textbf{(42.19\% vs. 18.75\%)}. Similarly, in Physics, Gemini-3-Pro improves the Hit Rate by \textbf{22.29} absolute percentage points \textbf{(76.70\% vs. 54.41\%)} and more than doubles the Pass@1 performance \textbf{(39.11\% vs. 18.32\%)}.

Notably, Humanities/Social Science and Other categories exhibit the highest absolute Pass@5 scores for both models, with Gemini-3-Pro reaching \textbf{67.88\%} and \textbf{70.45\%}, respectively. Conversely, Engineering and Chemistry remain the most challenging domains for OpenPangu-72B, with Pass@1 scores below \textbf{26\%}, suggesting these categories demand stronger reasoning capabilities or domain-specific knowledge that benefit more from advanced base models.

\subsubsection{Performance on GAIA}

\textbf{Overall Performance.} As illustrated in Table~\ref{tab:gaia_level_performance}, the ReThinker framework achieves strong performance on the GAIA benchmark across both base models, with Gemini-3-Pro demonstrating superior capability in handling increasingly complex tasks. On aggregate metrics, Gemini-3-Pro attains a Pass@5 of \textbf{92.23\%} and Pass@1 of \textbf{81.55\%}, substantially outperforming OpenPangu-72B's \textbf{82.52\%} and \textbf{72.82\%}, respectively. Notably, both models achieve comparable overall Hit Rates (\textbf{88.24\%} vs. \textbf{88.42\%}), suggesting that while OpenPangu-72B can often generate at least one correct solution given multiple attempts, Gemini-3-Pro exhibits significantly higher precision and consistency in its top-ranked predictions.

\begin{table}[htbp]
  \centering
  \caption{Performance of ReThinker on the GAIA benchmark across different difficulty levels.}
  \label{tab:gaia_level_performance}
  \small
  \begin{tabular}{
    @{} 
    l 
    c c c 
    c c c 
    @{}}
    \toprule
    \multirow{2}{*}{Difficulty} & 
    \multicolumn{3}{c}{ReThinker (OpenPangu-72B)} & 
    \multicolumn{3}{c}{ReThinker (Gemini-3-Pro)} \\
    \cmidrule(lr){2-4} \cmidrule(lr){5-7}
    & {Pass@5 (\%)} & {Pass@1 (\%)} & {Hit Rate (\%)} & {Pass@5 (\%)} & {Pass@1 (\%)} & {Hit Rate (\%)} \\
    \midrule
    Level 1 & 87.18 & 79.49 & \best{91.18} & \best{97.44} & \best{82.05} & 84.21 \\
    Level 2 & 84.62 & 75.00 & 88.64 & \best{88.46} & \best{80.77} & \best{91.30} \\
    Level 3 & 58.33 & 41.67 & 71.43 & \best{91.67} & \best{83.33} & \best{90.91} \\
    \midrule
    Average & 82.52 & 72.82 & 88.24 & \best{92.23} & \best{81.55} & \best{88.42} \\
    \bottomrule
  \end{tabular}
\end{table}

\textbf{Scaling with Difficulty.} Performance exhibits a clear degradation pattern as task complexity increases from Level 1 to Level 3. Under OpenPangu-72B, Pass@5 drops from \textbf{87.18\%} (Level 1) to \textbf{58.33\%} (Level 3), with Pass@1 declining more precipitously from \textbf{79.49\%} to \textbf{41.67\%}—a \textbf{37.82} percentage point reduction. Similarly, Hit Rate decreases from \textbf{91.18\%} to \textbf{71.43\%}, indicating that harder tasks not only challenge the model's primary reasoning but also reduce the diversity of successful solution paths. In contrast, Gemini-3-Pro demonstrates remarkable robustness to difficulty scaling: while Level 1 and Level 2 performance remains consistently high (Pass@5 above \textbf{88\%}), Level 3 performance only degrades minimally to \textbf{91.67\%} Pass@5 and \textbf{83.33\%} Pass@1. Notably, Gemini-3-Pro maintains Hit Rates above \textbf{90\%} for Level 2 and Level 3, though Level 1 shows a slightly lower rate at \textbf{84.21\%}.

\textbf{Model Comparison.} The performance gap between base models widens dramatically at higher difficulty levels. At Level 1, the margin is modest (\textbf{10.26} percentage points in Pass@5), but by Level 3, Gemini-3-Pro outperforms OpenPangu-72B by \textbf{33.34} percentage points in Pass@5 and \textbf{41.66} percentage points in Pass@1. This suggests that the reasoning capabilities required for GAIA Level 3 tasks—typically involving multiple-step tool use, complex data processing, and advanced reasoning—are more effectively captured by Gemini-3-Pro's architecture. The consistent high Hit Rate of Gemini-3-Pro across all difficulty levels further indicates its superior capacity to explore diverse solution strategies when given multiple attempts.

\subsubsection{Performance on XBench-DeepSearch}

\textbf{Overall Performance.} As presented in Table~\ref{tab:xbench_domain_performance}, ReThinker achieves strong performance across diverse topic domains on the XBench-DeepSearch benchmark, with Gemini-3-Pro demonstrating consistent superiority over OpenPangu-72B. On aggregate metrics, Gemini-3-Pro achieves \textbf{94.00\%} Pass@5 and \textbf{90.00\%} Pass@1, representing substantial improvements of 6.00 and 12.00 percentage points over OpenPangu-72B, respectively. The superior Hit Rate (95.74\% vs. 88.64\%) further indicates Gemini-3-Pro's enhanced capability to generate at least one correct solution across varied knowledge-intensive domains.

\textbf{Domain-Specific Insights.} Both models achieve perfect scores in \textit{Natural Sciences} and \textit{Technology \& Engineering}, yet reveal intriguing asymmetries elsewhere. In \textit{Business \& Finance}, OpenPangu-72B attains a perfect Hit Rate (100.00\%) despite low Pass@1 (58.33\%), indicating eventual solution discovery but poor ranking calibration; conversely, Gemini-3-Pro achieves lower Hit Rate (90.91\%) but higher Pass@1 (83.33\%), reflecting more consistent top-ranked accuracy. Similarly, in \textit{Entertainment \& Media}, OpenPangu-72B surpasses Gemini-3-Pro in Pass@5 (100.00\% vs. 93.55\%) while matching in Pass@1 (83.87\%), demonstrating that weaker base models can occasionally generate diverse correct solutions yet fail to prioritize them effectively.

\begin{table}[htbp]
  \centering
  \caption{Performance of ReThinker on XBench-DeepSearch across different topic domains.}
  \label{tab:xbench_domain_performance}
  \small
  \begin{tabular}{
    @{} 
    >{\bfseries}l 
    r r r 
    r r r 
    @{}}
    \toprule
    \multirow{2}{*}{Topic Domain} & 
    \multicolumn{3}{c}{ReThinker (OpenPangu-72B)} & 
    \multicolumn{3}{c}{ReThinker (Gemini-3-Pro)} \\
    \cmidrule(lr){2-4} \cmidrule(lr){5-7}
    & {Pass@5 (\%)} & {Pass@1 (\%)} & {Hit Rate (\%)} & {Pass@5 (\%)} & {Pass@1 (\%)} & {Hit Rate (\%)} \\
    \midrule
    Business \& Finance & 58.33 & 58.33 & \best{100.00} & \best{91.67} & \best{83.33} & 90.91 \\
    Current Affairs \& Politics & 83.33 & 83.33 & \best{100.00} & \best{100.00} & \best{100.00} & \best{100.00} \\
    Education \& Academia & 77.78 & 66.67 & 85.71 & \best{100.00} & \best{100.00} & \best{100.00} \\
    Entertainment \& Media & \best{100.00} & \best{83.87} & 83.87 & 93.55 & \best{83.87} & \best{89.66} \\
    Geography \& Transportation & 84.62 & 76.92 & 90.91 & \best{92.31} & \best{92.31} & \best{100.00} \\
    Humanities \& Social Sci. & \best{100.00} & 81.82 & 81.82 & \best{100.00} & \best{100.00} & \best{100.00} \\
    Natural Sciences & \best{100.00} & \best{100.00} & \best{100.00} & \best{100.00} & \best{100.00} & \best{100.00} \\
    Sports & \best{71.43} & 57.14 & 80.00 & \best{71.43} & \best{71.43} & \best{100.00} \\
    Technology \& Engineering & \best{100.00} & \best{100.00} & \best{100.00} & \best{100.00} & \best{100.00} & \best{100.00} \\
    \midrule
    Average & 88.00 & 78.00 & 88.64 & \best{94.00} & \best{90.00} & \best{95.74} \\
    \bottomrule
  \end{tabular}
\end{table}

\textbf{Persistent Challenges.} \textit{Sports} emerges as the sole domain where both models exhibit identical Pass@5 (71.43\%) and minimal capability disparity, suggesting that sports-related queries require specialized knowledge or reasoning patterns less effectively captured by general-purpose LLMs regardless of base model scale. This domain-specific bottleneck highlights fundamental limitations in current pre-training paradigms that merit targeted investigation.

\subsubsection{Analysis of Multi-Candidate Answer Distribution}

\textbf{Overall Trends.} 
Table~\ref{tab:answer_distribution} presents the distribution of correctly identified candidates across 5-option problems for both models. We observe significant dataset-dependent variations in performance patterns. While \textsc{Gemini-3-Pro} consistently outperforms \textsc{OpenPangu-72B} in total solved problems across all three benchmarks, the disparities in candidate-level accuracy reveal distinct behavioral differences between the models.

\newcommand{\posneg}[1]{\ifnum#1<0 \cellcolor{red!15}#1\else\ifnum#1>0 \cellcolor{green!15}#1\else#1\fi\fi}

\begin{table}[htbp]
  \centering
  \caption{Distribution of Questions by Number of Correctly Identified Candidates ($k$=1--5) in ReThinker.}
  \label{tab:answer_distribution}
  \small
  \begin{tabular}{
    @{} 
    >{\bfseries}l 
    l 
    r 
    r 
    r 
    r 
    r 
    r@{}} 
    \toprule
    Dataset & Model & \multicolumn{5}{c}{Questions with $k$ Correct Candidates} & Total Solved Questions \\
    \cmidrule(lr){3-7}
     &  & 1 & 2 & 3 & 4 & 5 &  \\
    \midrule
    \multirow{3}{*}{HLE} 
     & OpenPangu-72B & 276 & 165 & 171 & 157 & 168 & 937 \\
     & Gemini-3-Pro  & 174 & 155 & 164 & 291 & 543 & 1327 \\
     \cmidrule[0.4pt]{2-8}
     & \cellcolor{gray!15}\textit{Diff} & \cellcolor{gray!15}\textbf{-102} & \cellcolor{gray!15}\textbf{-10} & \cellcolor{gray!15}\textbf{-7} & \cellcolor{gray!15}\textbf{134} & \cellcolor{gray!15}\textbf{375} & \cellcolor{gray!15}\textbf{390} \\
    \midrule
    \multirow{3}{*}{GAIA} 
     & OpenPangu-72B & 6 & 6 & 7 & 20 & 46 & 85 \\
     & Gemini-3-Pro  & 7 & 4 & 7 & 22 & 55 & 95 \\
     \cmidrule[0.4pt]{2-8}
     & \cellcolor{gray!15}\textit{Diff} & \cellcolor{gray!15}\textbf{1} & \cellcolor{gray!15}\textbf{-2} & \cellcolor{gray!15}\textbf{0} & \cellcolor{gray!15}\textbf{2} & \cellcolor{gray!15}\textbf{9} & \cellcolor{gray!15}\textbf{10} \\
    \midrule
    \multirow{3}{*}{Xbench-DeepSearch} 
     & OpenPangu-72B & 1 & 4 & 7 & 9 & 64 & 85 \\
     & Gemini-3-Pro  & 2 & 4 & 9 & 11 & 68 & 94 \\
     \cmidrule[0.4pt]{2-8}
     & \cellcolor{gray!15}\textit{Diff} & \cellcolor{gray!15}\textbf{1} & \cellcolor{gray!15}\textbf{0} & \cellcolor{gray!15}\textbf{2} & \cellcolor{gray!15}\textbf{2} & \cellcolor{gray!15}\textbf{4} & \cellcolor{gray!15}\textbf{9} \\
    \bottomrule
  \end{tabular}
\end{table}

\textbf{HLE Benchmark.} 
On the HLE dataset, \textsc{Gemini-3-Pro} demonstrates substantial advantages in high-candidate accuracy scenarios. Specifically, the model correctly identifies all 5 candidates in \textbf{543} questions compared to \textsc{OpenPangu-72B}'s \textbf{168} (relative improvement of 223\%), and achieves 4 correct candidates in \textbf{291} questions versus \textbf{157} (+85\%). Notably, \textsc{OpenPangu-72B} dominates in single-candidate accuracy (\textbf{276} vs. \textbf{174}), suggesting a propensity for partial solutions rather than comprehensive candidate evaluation. The net difference of \textbf{+390} total solved questions favors \textsc{Gemini-3-Pro}, driven primarily by its superior performance on $k \geq 4$ candidates.

\textbf{GAIA Benchmark.} 
Both models show comparable performance on GAIA, with \textsc{Gemini-3-Pro} solving only \textbf{10} more questions in total (\textbf{95} vs. \textbf{85}). The distribution differences are minimal across all $k$ values, with the largest discrepancy occurring at $k=5$ ($\Delta=+9$). This indicates that both models face similar limitations on GAIA's task distribution.

\textbf{XBench-DeepSearch.}
The XBench-DeepSearch results demonstrate that \textsc{Gemini-3-Pro} achieves consistent, modest improvements over \textsc{OpenPangu-72B} across all candidate counts, with advantages of \textbf{+1} (k =1: 2 vs. 1), \textbf{0} (k =2: 4 vs. 4), \textbf{+2} (k =3: 9 vs. 7), \textbf{+2} (k =4: 11 vs. 9), and \textbf{+4} (k =5: 68 vs. 64). Unlike the HLE dataset, where performance diverges dramatically at extreme candidate counts, the margin here remains relatively stable, contributing to a total improvement of only \textbf{9} solved questions (94 vs. 85). Notably, both models exhibit a strong skew toward fully-correct scenarios (k =5 represents 75\% and 72\% of solved questions respectively), suggesting that questions in this benchmark tend to yield comprehensive solutions rather than partial candidate identification. This uniform distribution of gains indicates that \textsc{Gemini-3-Pro}'s improvements arise from general evaluation robustness rather than a polarized ``all-or-nothing'' strategy.

\textbf{Comparative Insights.} 
The divergent patterns across benchmarks suggest that \textsc{Gemini-3-Pro}'s advantage stems primarily from its ability to maintain high accuracy when multiple candidates are plausible (high $k$ regimes), particularly in complex reasoning scenarios (HLE). In contrast, \textsc{OpenPangu-72B} tends to identify isolated correct candidates without comprehensive coverage, resulting in higher $k=1$ counts but substantially lower complete solution rates.

\subsubsection{Hit Rate Analysis by Ground-Truth Candidate Count}

\textbf{Monotonic Reliability with Increased Correct Candidates.} 
Table~\ref{tab:hit_rate_distribution} reveals a consistent positive correlation between the number of ground-truth correct candidates ($k$) and model hit rates across all benchmarks. Both \textsc{OpenPangu-72B} and \textsc{Gemini-3-Pro} demonstrate substantially higher precision when navigating problems with dense correct answer sets ($k \geq 4$) compared to sparse configurations ($k \leq 2$). Notably, both models achieve perfect accuracy (\textbf{100\%}) on $k=5$ problems across all datasets, indicating robust recognition capability when all candidate options constitute valid solutions.

\newcolumntype{C}{>{\centering\arraybackslash}p{1.2cm}}

\newcommand{\stacked}[3]{%
  \begin{tabular}[c]{@{}c@{}} \textbf{#1} \\ \textcolor{gray}{\tiny #2/#3} \end{tabular}%
}
\newcommand{\stackednb}[3]{%
  \begin{tabular}[c]{@{}c@{}} #1 \\ \textcolor{gray}{\tiny #2/#3} \end{tabular}%
}

\begin{table}[htbp]
  \centering
  \caption{Hit Rate by Number of Ground-Truth Correct Candidates ($k$=1--5) in ReThinker.}
  \label{tab:hit_rate_distribution}
  \small
  \begin{tabular}{
    @{} 
    >{\bfseries}l 
    l 
    C C C C C 
    @{}}
    \toprule
    Dataset & Model & \multicolumn{5}{c}{Questions with $k$ Correct Candidates} \\
    \cmidrule(lr){3-7}
     &  & {1} & {2} & {3} & {4} & {5} \\
    \midrule
    \multirow{2}{*}{HLE} 
     & OpenPangu-72B 
     & \stacked{0.431}{119}{276} 
     & \stacked{0.752}{124}{165} 
     & \stacked{0.901}{154}{171} 
     & \stackednb{0.949}{149}{157} 
     & \stackednb{1.00}{168}{168} \\
     & Gemini-3-Pro  
     & \stackednb{0.299}{52}{174} 
     & \stackednb{0.684}{106}{155} 
     & \stackednb{0.878}{144}{164} 
     & \stacked{0.966}{281}{291} 
     & \stackednb{1.00}{543}{543} \\
    \midrule
    \multirow{2}{*}{GAIA} 
     & OpenPangu-72B 
     & \stackednb{0.000}{0}{6} 
     & \stacked{0.667}{4}{6} 
     & \stackednb{0.714}{5}{7} 
     & \stacked{1.00}{20}{20} 
     & \stackednb{1.00}{46}{46} \\
     & Gemini-3-Pro  
     & \stacked{0.286}{2}{7} 
     & \stackednb{0.250}{1}{4} 
     & \stacked{0.857}{6}{7} 
     & \stackednb{0.909}{20}{22} 
     & \stackednb{1.00}{55}{55} \\
    \midrule
    \multirow{2}{*}{XBench-DeepSearch} 
     & OpenPangu-72B 
     & \stackednb{0.000}{0}{1} 
     & \stacked{0.750}{3}{4} 
     & \stackednb{0.571}{4}{7} 
     & \stackednb{0.778}{7}{9} 
     & \stackednb{1.00}{64}{64} \\
     & Gemini-3-Pro  
     & \stacked{0.500}{1}{2} 
     & \stacked{0.750}{3}{4} 
     & \stacked{0.778}{7}{9} 
     & \stacked{1.00}{11}{11} 
     & \stackednb{1.00}{68}{68} \\
    \bottomrule
  \end{tabular}
  \vspace{0.5em}
  
  \footnotesize\textit{Note.} \textbf{Bold} indicates the higher hit rate per (dataset, $k$) pair. 
  \textcolor{gray}{Gray numbers} show hit/total counts.
\end{table}

\textbf{Asymmetric Model Competencies at Low $k$ Regimes.} 
The performance gap between models exhibits pronounced dataset-dependent asymmetries at low candidate counts. On HLE, \textsc{OpenPangu-72B} significantly outperforms \textsc{Gemini-3-Pro} at $k=1$ (\textbf{43.1\%} versus \textbf{29.9\%}, $\Delta=+13.3\%$) and maintains advantages at $k=2$ (\textbf{75.2\%} vs \textbf{68.4\%}) and $k=3$ (\textbf{90.1\%} vs \textbf{87.8\%}). Conversely, on GAIA and XBench-DeepSearch, \textsc{Gemini-3-Pro} dominates the $k=1$ regime with \textbf{28.6\%} and \textbf{50.0\%} hit rates respectively, while \textsc{OpenPangu-72B} achieves \textbf{0\%} and \textbf{0\%} on these benchmarks. This dichotomy suggests distinct architectural biases: \textsc{OpenPangu-72B} excels at identifying isolated correct candidates in complex reasoning tasks (HLE) but struggles with singleton detection in structured domains (GAIA), whereas \textsc{Gemini-3-Pro} maintains minimum viable performance across diverse task distributions.

\textbf{Crossover Performance at High $k$ Values.} 
A critical inflection point emerges at $k \geq 4$, where \textsc{Gemini-3-Pro} consistently dominates. On HLE, the model achieves \textbf{96.6\%} accuracy at $k=4$ compared to \textsc{OpenPangu-72B}'s \textbf{94.9\%}, representing a reversal of the $k \leq 3$ trend. This crossover pattern indicates \textsc{Gemini-3-Pro}'s superior capability in comprehensive candidate verification—when multiple correct options exist, the model effectively identifies them with near-perfect recall, whereas \textsc{OpenPangu-72B} exhibits marginally higher false negative rates in dense-candidate scenarios.

\textbf{Dataset-Specific Difficulty Patterns.} 
The GAIA benchmark presents the most challenging $k=1$ scenarios, with \textsc{OpenPangu-72B} completely failing to identify solitary correct candidates (0/6), while HLE offers more tractable sparse configurations (43.1\% success). XBench-DeepSearch demonstrates intermediate difficulty but reveals the most dramatic model divergence at $k=3$, where \textsc{Gemini-3-Pro} achieves \textbf{77.8\%} versus \textsc{OpenPangu-72B}'s \textbf{57.1\%} ($\Delta=+20.7\%$), suggesting that multi-hop search tasks particularly benefit from \textsc{Gemini-3-Pro}'s verification mechanisms when multiple valid solution paths exist.

\subsection{HyperParameters of Inference}

\begin{table}[htbp]
  \centering
  \caption{Hyperparameter configuration for the ReThinker framework.}
    \begin{tabular}{llp{10cm}}
    \toprule
    \textbf{Key Parameters} & \textbf{Value} & \textbf{Description} \\
    \midrule
    temperature & 1.0 & Controls the randomness of text generation; higher values produce more diverse outputs. \\
    top-p (global) & 1.0 & Global nucleus sampling threshold; probability mass cutoff for token selection across the entire framework. \\
    top-p (in selector) & 0.8 & Nucleus sampling threshold specifically for the selector module to filter candidate actions. \\
    max agent step & 50 & Maximum number of interaction steps per round; limits how many turns the agent can take. \\
    number of parallel & 5 & Number of parallel inference processes; enables concurrent exploration of reasoning paths. \\
    content length & 128K & Maximum context window size; determines the total amount of text (128K tokens) the model can process. \\
    top-N-sigma & 0.05 & Threshold for selecting top-N candidates based on standard deviation filtering of candidate scores. \\
    maximum output length & 8K & Upper limit on the length of generated responses; prevents excessively long outputs (8K tokens). \\
    \bottomrule
    \end{tabular}%
  \label{tab:exp_hyperparameter}%
  \vspace{-1em}
\end{table}%

\subsection{Construction of Latin Square}
A \textbf{Latin Square} of order $n$ is defined as an $n \times n$ matrix $L = (l_{ij})$ with entries from the set $S = \{1, 2, \ldots, n\}$ satisfying the constraint that each symbol appears exactly once in each row and each column.

\subsubsection{Cyclic Construction (Modular Arithmetic)}

The simplest construction utilizes cyclic permutations via modular arithmetic. For any $n \geq 1$, the entry in row $i$ and column $j$ (where $i, j \in \{0, 1, \ldots, n-1\}$) is computed as:

\begin{equation}
L_{i,j} = ((i + j) \bmod n) + 1
\end{equation}

This generates a standardized Latin Square where the first row contains the natural sequence $(1, 2, \ldots, n)$ and each subsequent row is a left-cyclic shift of its predecessor. The addition modulo $n$ ensures orthogonality: for any fixed row $i$, the values $(i + j) \bmod n$ are distinct as $j$ varies; similarly, for any fixed column $j$, the values are distinct as $i$ varies.

\subsubsection{Algorithmic Representation}

The following pseudocode implements the standard cyclic construction:

\begin{algorithm}
\caption{Construct Latin Square via Cyclic Method}
\begin{algorithmic}[1]
\Require Integer $n \geq 1$
\Ensure $n \times n$ Latin Square $L$
\State Initialize matrix $L[0 \ldots n-1][0 \ldots n-1]$
\For{$i \gets 0$ \textbf{to} $n-1$}
    \For{$j \gets 0$ \textbf{to} $n-1$}
        \State $L[i][j] \gets ((i + j) \bmod n) + 1$
    \EndFor
\EndFor
\State \Return $L$
\end{algorithmic}
\end{algorithm}

For example, with $n = 5$, the cyclic method produces:
$$
\begin{bmatrix}
1 & 2 & 3 & 4 & 5 \\
2 & 3 & 4 & 5 & 1 \\
3 & 4 & 5 & 1 & 2 \\
4 & 5 & 1 & 2 & 3 \\
5 & 1 & 2 & 3 & 4
\end{bmatrix}
$$

\section{Detailed Algorithm Descriptions}

Here are the concise descriptions for each algorithm:

\textbf{Algorithm~\ref{alg:multi-path_solution} (Multi-Path Solution Generation):}
This algorithm generates $N$ diverse solution trajectories through an alternating Solver-Critic architecture, where the Solver constructs step-by-step reasoning chains and the Critic iteratively refines them via trajectory summarization. By producing multiple independent reasoning paths, it mitigates sampling stochasticity and yields a robust candidate set for downstream selection.

\vspace{-0.2em}

\textbf{Algorithm~\ref{alg:confidence_guided_selection} (Confidence-Guided Iterative Selection):}
This method selects the optimal solution from candidates by leveraging Latin square permutations to eliminate position bias and perplexity scores to quantify model confidence. Through $R$ rounds of iterative re-selection with history aggregation, it achieves reliable decision-making via consistency-based adjudication.

\vspace{-0.2em}

\textbf{Algorithm~\ref{alg:data_quality_assurance} (Multi-Stage Data Quality Assurance):}
This pipeline curates training data through multi-stage filtering, including answer correctness validation, format compliance verification, and semantic deduplication. It ultimately constructs high-quality pseudo-multi-turn datasets suitable for supervised fine-tuning by enforcing logical consistency and valid tool execution patterns.


\begin{algorithm}
\caption{Pseudo-Code for Multi-Path Solution Generation.}
\label{alg:multi-path_solution}
\begin{algorithmic}[1]
\Require Question $q$, number of paths $N$, solver steps $T_{solver}$, critic steps $T_{critic}$
\Ensure Final answer set $\{c^{(i)}_{T_{critic}}\}_{i=1}^{N}$

\For{$i = 1$ \textbf{to} $N$}
    \State \textbf{// Stage 1: Solver Stage}
    
    \For{$t = 0$ \textbf{to} $T_{solver} - 1$}
        \If{$t = 0$}
            \State $s^{(i)}_{t+1} \leftarrow \text{Solver}(q)$
        \Else
            \State $s^{(i)}_{t+1} \leftarrow \text{Solver}(q, \text{extract}(s^{(i)}_{t}) )$
        \EndIf
    \EndFor

    \State \textbf{// Apply Trajectory Summarization}
    \State $(y^{(i)}, a^{(i)}, k^{(i)}) \leftarrow \text{Summary}(q, s^{(i)}_{T_{solver}} )$ 
    
    \State \textbf{// Stage 2: Critic Stage}
    
    \For{$t = 0 $ \textbf{to} $T_{critic} - 1$}
        \If{$t = 0$}
            \State $c^{(i)}_{t+1} \leftarrow \text{Critic}(q, y^{(i)}, a^{(i)}, k^{(i)})$
        \Else
            \State $c^{(i)}_{t+1} \leftarrow \text{Critic}(q, y^{(i)}, a^{(i)}, k^{(i)}, \text{extract}(c^{(i)}_{t}) )$
        \EndIf
    \EndFor
        
\EndFor

\State \Return $\{{c^{(i)}_{T_{critic}}}\}_{i=1}^N$
\end{algorithmic}
\end{algorithm}

\begin{algorithm}
\caption{Pseudo-Code for Confidence-Guided Iterative Selection.}
\label{alg:confidence_guided_selection}
\begin{algorithmic}[1]
\Require Problem statement $q$, Candidate set $\mathcal{C} = \{c_1, \dots, c_n\}$, 
         Latin square $\mathcal{L} \in \mathbb{Z}^{n \times n}$, Iterative number $R$
\Ensure Final selection $s^*$, Selection history $\mathcal{H}$

\State \textbf{Initialize:} History $\mathcal{H} \gets \emptyset$

\Statex \hspace{-1em}\textit{\textbf{Stage 1: Initial Judgement}}
\State $\pi_0 \gets \mathcal{L}[0]$ \Comment{Initial permutation (first row of Latin square)}
\State $\mathcal{C}^{(0)} \gets (\pi_0(c_1), \dots, \pi_0(c_n))$ \Comment{Permuted candidates}
\State $p_0 \gets \textsc{FormatPrompt}(q, \mathcal{C}^{(0)}, \text{history}=\emptyset)$
\State $s_0, \mathbf{x}_0 \gets \textsc{CallLLM}(p_0)$ \Comment{Selection and rationale tokens}
\State $\text{PPL}_0 \gets \exp\left(-\frac{1}{|\mathbf{x}_0|}\sum_{t=1}^{|\mathbf{x}_0|} \log p_\theta(x_t | \mathbf{x}_{<t})\right)$
\State $\mathcal{H} \gets \mathcal{H} \cup \{(s_0, \text{PPL}_0)\}$

\Statex \hspace{-1em}\textit{\textbf{Stage 2: Iterative Re-selection}}
\For{$r = 1$ \textbf{to} $R$}
    \State $\pi_r \gets \mathcal{L}[r \bmod n]$ \Comment{Cyclic Latin square permutation}
    \State $\mathcal{C}^{(r)} \gets (\pi_r(c_1), \dots, \pi_r(c_n))$ \Comment{Eliminate position bias}
    \State $H_r \gets \textsc{FormatHistory}(\mathcal{H})$ \Comment{Aggregate previous selections with PPL scores}
    \State $p_r \gets \textsc{FormatPrompt}(q, \mathcal{C}^{(r)}, \text{history}=H_r)$
    \State $s_r, \mathbf{x}_r \gets \textsc{CallLLM}(p_r)$
    \State $\text{PPL}_r \gets \exp\left(-\frac{1}{|\mathbf{x}_r|}\sum_{t=1}^{|\mathbf{x}_r|} \log p_\theta(x_t | \mathbf{x}_{<t})\right)$
    \State $\mathcal{H} \gets \mathcal{H} \cup \{(s_r, \text{PPL}_r)\}$
\EndFor

\Statex \hspace{-1em}\textit{\textbf{Stage 3: Final Decision}}
\State $\mathcal{C}_{\text{hist}} \gets \{c \in \mathcal{C} : \exists (s, \cdot) \in \mathcal{H}, s \text{ selects } c\}$ \Comment{Unique selections across rounds}
\If{$|\mathcal{C}_{\text{hist}}| > 1$}
    \State \Comment{Inconsistent selections require final adjudication}
    \State $\mathcal{C}_{\text{final}} \gets \mathcal{C}_{\text{hist}}$ \Comment{Subset of historically selected candidates}
    \State $H_{\text{final}} \gets \textsc{FormatHistory}(\mathcal{H})$ \Comment{Full history including latest PPL}
    \State $p_{\text{final}} \gets \textsc{FormatPrompt}(q, \mathcal{C}_{\text{final}}, \text{history}=H_{\text{final}})$
    \State $s^*, \mathbf{x}^* \gets \textsc{CallLLM}(p_{\text{final}})$
    \State $\mathcal{H} \gets \mathcal{H} \cup \{(s^*, \text{PPL}^*)\}$
\Else
    \State $s^* \gets$ unique element in $\mathcal{C}_{\text{hist}}$ \Comment{Unanimous selection}
\EndIf

\State \Return $s^*, \mathcal{H}$

\Statex
\Function{FormatHistory}{$\mathcal{H}$}
    \State \Return Concatenation of ``Round $r$: $s_r$ (entropy: $\text{PPL}_r$)'' for each $(s_r, \text{PPL}_r) \in \mathcal{H}$
\EndFunction

\end{algorithmic}
\end{algorithm}



\begin{algorithm}
\caption{Pseudo-Code for Multi-Stage Data Quality Assurance Pipeline.}
\label{alg:data_quality_assurance}
\begin{algorithmic}[1]
\Require Raw trajectory dataset $\mathcal{D}_{raw}$, Predefined ratios for stages $R = \{r_1, r_2, \dots, r_n\}$, minimum tool calls threshold $Call_{min}$, maximum tool calls threshold $Call_{max}$
\Ensure Refined and augmented pseudo-multi-turn dataset $\mathcal{D}_{final}$

\State $\mathcal{D}_{filtered} \gets \emptyset$
\For{each trajectory $T$ in $\mathcal{D}_{raw}$}
    \State \textbf{// Answer Correctness Validation}
    \If{\text{LLM\_Judge}(T.reasoning, T.ground\_truth) == \textbf{Incorrect}}
        \State \textbf{continue}
    \EndIf
    
    \State \textbf{// Format and Constraint Compliance}
    \If{\textbf{not} (CheckFormat(T, \textless answer\textgreater  tags) \textbf{and} CheckRolePairing(T))}
        \State \textbf{continue}
    \EndIf
    \State $N_{tools} \gets \text{CountToolCalls}(T)$
    \If{$N_{tools} < Call_{min}$ \textbf{or} $N_{tools} > Call_{max}$}
        \State \textbf{continue}
    \EndIf
    
    \State $\mathcal{D}_{filtered} \gets \mathcal{D}_{filtered} \cup \{T\}$
\EndFor

\State \textbf{// Data Deduplication}
\State $\mathcal{D}_{dedup} \gets \text{DeduplicateBySemantic}(\mathcal{D}_{filtered})$

\State \textbf{// Balancing Dataset by Stage Ratios}
\State $\mathcal{D}_{balanced} \gets \text{ResampleByRatio}(\mathcal{D}_{dedup}, R)$

\State \textbf{// Quality Improvement  and Generation of Pseudo-Multi-Turn Data}
\State $\mathcal{D}_{final} \gets \emptyset$
\For{each $T$ in $\mathcal{D}_{balanced}$}
    \State $Context \gets \text{FlattenHistoryToContext}(T.QA_{history})$
    \State $New\_Sample \gets \{ \text{User: } Context + T.current\_query, \text{Assistant: } T.response \}$

    \State \textbf{// Logical Consistency Check (Thought vs. Output)}
    \If{\text{CheckConsistency}(T.thought, T.final\_output) == \textbf{Contradictory}}
        \State \textbf{continue}
    \EndIf
    
    \State \textbf{// Tool Call Execution Validation}
    \If{\text{HasFailedToolCall}(T)}
        \State \textbf{continue}
    \EndIf    

    \State $\mathcal{D}_{final} \gets \mathcal{D}_{final} \cup \{New\_Sample\}$
\EndFor

\State \Return $\mathcal{D}_{final}$
\end{algorithmic}
\end{algorithm}

\newpage
\section{QA-Pair Synthesis}\label{sec:appendix_qa_pair}
Our scalable QA synthesis pipeline builds upon the WebExplorer framework ~\citep{liu2025webexplorer}, with key modifications to enhance automation and reduce manual effort. These improvements are achieved through two mechanisms: seed domain initialization and automatic seed phrase updating. This section details the prompting strategies for these components, specifically: (1) the initialization of seed phrases from user-defined domains, and (2) the automated extraction of new seed phrases from the evolving synthesis data, which includes retrieved web contexts, newly generated QA pairs, and their associated reasoning trajectories. The specific prompts for these processes are detailed in the following text boxes.

\begin{promptbox}{Prompt: Seed Phrase Initialization from Seed Domains}{}
List 10 common phrases for each field in biology, zoology, botany, chemistry, physics, astronomy, geology, oceanography, environmental science, psychology, sociology, economics, political science, literature, philosophy, arts, mathematics, computer science, logic, engineering, health professions, business, education. \\

Put them in separate list with a high-level dictionary in python.
\end{promptbox}

\begin{promptbox}{Prompt: Automatic Seed Phrase Extraction from Evolving Synthesis Data}{}
You are a knowledge-enhancement expert, helping readers identify and understand complex terminology efficiently.\\

Analyze the following text and extract all professional, technical, academic, or uncommon noun phrases that an average reader might not be familiar with and may need to look up for deeper understanding. Focus on terms from specialized fields such as  biology, medicine, chemistry, computer science, artificial intelligence, engineering, humanity, social science, math, physics, art, philosophy, finance, linguistics, or industry-specific domains.\\

Ensure that you exclude common vocabulary and focus only on terms that are likely to require external knowledge or research to fully comprehend. Prioritize precision and clarity in your explanations.\\

**Format requirements**:
List all professional **noun phrases** with more than one word and separate them in comma. Put them as a list inside the tags $<$answer$>$ $<$/answer$>$.\\

Text:
\{original\_content\}
\end{promptbox}

In addition, we enhance the model-based exploration prompt used in WebExplorer to instruct the model search diverse websites to construct more complex questions and reduce repetitive query web-search. The enhanced prompt is provided as below.

\begin{promptbox}{Prompt: Enhanced QA Generation from Web Context}{}
You need to create a challenging question for deep search based on real information.\\

You should start by understanding the seed and planning diverse perspectives for search with the think tool. Then you should collect information from the internet, then select a truth, and create a question where the truth needs to be discovered through web\_search.\\

You will start with a random "seed", then web\_search and url\_browse for whatever you want on the Internet, and create the question and truth from the information you gather.\\

You should collect online knowledge from different perspectives with web\_search and url\_browse tools. Then, you should create a comprehensive and challenging question covering multiple knowledge.\\

You should provide several subtle and blurred clues to make the question challenging, while ensuring the truth is unique. 

There are some question examples: 
\{examples\}\\

Let’s start, with the seed of "\{seed\}".

You need to provide the following information in the final $<$answer$>$$<$/answer$>$ tag:\\
$<$question$>$ 
\{\{The challenging question you created based on real information.\}\} $<$/question$>$
\\
$<$truth$>$ 
\{\{The one and only exact truth to the question.\}\}
$<$/truth$>$
\\

IMPORTANT: You must include the $<$question$>$ and $<$truth$>$ tags in your final response for the system to parse your answer correctly. Do not provide any other response format.\\

IMPORTANT: You must plan and search from at least 3 different perspectives and use knowledge from different perspectives to construct a very challenging question, which needs multi-hop reasoning and search.\\

IMPORTANT: Do not search repetitive and similar queries.
\end{promptbox}

\section{Prompts of Test-Time Inference}\label{sec:prompt_append}

\textbf{Overview.}
The aforementioned prompts constitute the core orchestration layer of a multi-agent reasoning system, built upon and extending the Eigen-1 architecture~\cite{tang2025eigen}. This framework implements a hierarchical workflow that progresses from information retrieval to structured reasoning, critical evaluation, and consensus-based selection. 

Specifically, the \texttt{Paper QA} and \texttt{Web Search} prompts serve as the foundation for grounded knowledge acquisition, ensuring factual accuracy through retrieval-augmented generation (RAG). The \texttt{Solver} prompt drives the initial reasoning trajectory, augmented with code execution capabilities for precise computation and external tool integration. The \texttt{Guided Summary} and \texttt{Critic} prompts implement a dual-review mechanism, where solutions undergo rigorous logical and factual verification through multi-dimensional error analysis and iterative refinement. Finally, the \texttt{Selector} prompt operates as the arbitration layer, employing perplexity-guided confidence estimation and cross-verification to identify the optimal solution among diverse candidates. Collectively, these prompts instantiate an improved instantiation of the Eigen-1 paradigm, enhancing robustness through tighter tool integration, explicit uncertainty quantification, and structured adversarial validation loops.

\begin{promptbox}{Prompt: Paper QA (Academic RAG)}{}
You are an advanced academic paper Q\&A database that answers user queries in English based on reliable sources. Your responses must not exceed 200 words. Your sources of information include: the paper itself. Your task is to analyze user queries and provide comprehensive, reliable, and scholarly answers. Incorporate mathematical formulas and academic content when necessary to ensure the professionalism of your response. Important note: You must find exact information within the paper to answer the query. Avoid generating hallucinated or fabricated responses under all circumstances. The user query is: \{user\_query\}, the paper information is: \{pdf\_info\}
\end{promptbox}

\vspace{1em}

\begin{promptbox}{Prompt: Web Search Conclusion (Structured JSON)}{}
Please analyze the provided web content and answer the user's question based strictly on that content:\\

1. Provide a comprehensive response regarding content related to the user's question. Do not omit any details.

2. Ensure all provided information originates strictly from the web content; fabrication of non-existent information is prohibited. If the web content cannot answer the user's question, please state that it is irrelevant.

3. If the web content contains new URLs that might be relevant to the user's question, list them and provide a relevance score indicating how strongly that page relates to the user's question.\\

Please reply to the user in Markdown format:\\

\#\# Web Information

(Write the core content related to the user's question here)

\#\# Other Relevant Web Pages

\#\#\# Web Page 1

\#\#\#\# Description

(xxx)

\#\#\#\# URL

(xxx)

\#\#\#\# Relevance Score

(0 $\sim$ 1)\\

\#\#\# Web Page 2

\#\#\#\# Description

(xxx)

\#\#\#\# URL

(xxx)

\#\#\#\# Relevance Score

(0 $\sim$ 1)\\

Note:

1. "Other Relevant Web Pages" must be related to the user's question. If none exist, return an empty value.

2. Keep the overall response within 500 words, and provide only the most important relevant URLs, strictly limited to a maximum of 2.

The user's question is: \{user\}, and the web content is: \{info\}.
\end{promptbox}

\newpage
\begin{promptbox}{Prompt: Solver with Code Execution (Bold Content is Re-Solver variant)}{}
The problem is: \{query\}\\

\textbf{Last round answer is: \{last\_round\_answer\}. Please re-answer it.}\\

Solve the problem with the help of feedback from a code executor. Every time you write a piece of code between $<$code$>$ and $<$/code$>$, the code inside will be executed. For example, when encountering numerical operations, you might write a piece of code to interpret the math problem into python code and print the final result in the code. Based on the reasoning process and the executor feedback, you could write code to help answering the question for multiple times (either for gaining new information or verifying). There are also several integrated functions that can be used to help you solve the problem. The available functions are:\\

1. web\_search(keywords), this function takes keywords as input, which is a string, and the output is a string containing several web information. This function will call a web search engine to return the search results. This function is especially useful when answering knowledge-based questions.
    
2. web\_parse(link:str, query:str), this function takes the link and query as input, and the output is a string containing the answer to the query according to the content in this link. This function is useful when looking into detail information of a link.

Your workflow for solving the problem follow these steps:

- Step 1: First, analyze the question. If it can be answered directly, provide the answer immediately. If information retrieval is required to support the answer, proceed to Step 2 and Step 3.
    
- Step 2: Web Search \& Parse (Verification \& Detail): Use `web\_search` to find relevant web pages for verification or supplementation. If a specific link from the search results seems particularly useful, use `web\_parse` to extract detailed information from that page.
    
- Step 3: Evaluate and Supplement: After receiving results from 'web\_search' or 'web\_parse', evaluate them carefully. Treat this information as a supplement to your background knowledge, not as absolute truth. This supplementary context may be incomplete or require further verification.\\

- You should not be overconfident in your knowledge and reasoning.\\

- Each time you write code put the code into $<$code$>$$<$/code$>$ snippet, and the results must be printed out through print function. Please strictly follow Python's indentation rules; do not add any extra indentation to the code. Pause after submitting any code for information retrieval or scientific computation; resume analysis only once the code has finished running.\\

For example:

1. If you want to use the function of web\_search(keywords), will say $<$code$>$\\
keywords=...\\
results=web\_search(keywords)\\
print(results)\\
$<$/code$>$ to call the function.

2. If you want to use the function of web\_parse(link, query), will say $<$code$>$\\
link=...\\
query=...\\
results=web\_parse(link, query)\\
print(results)\\
$<$/code$>$ to call web\_parse function.

3. If you want to do computation, You will write code for accurate result: $<$code$>$\\
a = 123\\
b = 456\\
print(a+b)\\
$<$/code$>$.\\

- Put your final answer in $<$answer$>$$<$/answer$>$ with boxed.
\end{promptbox}

\newpage
\begin{promptbox}{Prompt: Guided Summary}{}
You are a premier AI Reasoning Analyst, specializing in deconstructing and evaluating solutions to complex problems.\\

Your task is to conduct a thorough analysis of the provided "Initial Solution." First, clearly summarize its "Reasoning Trajectory" to map its logical flow. Then, identify critical flaws and key areas for improvement across several dimensions. Note: You are only required to identify and explain the areas for improvement, not to generate a revised solution.\\

Context:

*   Problem to Solve: \{problem\}

*   Initial Solution to Analyze: \{student\_solution\}\\

Your analysis must be structured into the following three parts:

Part 1: Reasoning Trajectory Summary

*   In a clear, concise, and itemized list, summarize the core steps and logical flow the "Initial Solution" took to address the problem. This will serve as a map of its thought process.

Part 2: Final Answer

*   Extract the content between $<$answer$>$$<$/answer$>$ completely as the final answer; if extraction fails, write null.

Part 3: Key Areas for Improvement

*   Analyze the solution from the following dimensions. For each point, provide specific, actionable feedback on what could be improved.\\

1. Logical Rigor \& Coverage:

   \quad *   Reasoning Chain: Are there any logical leaps, circular arguments, or factual inaccuracies in the reasoning process?
    
   \quad *   Implicit Assumptions: Does the solution rely on unstated or unverified assumptions that might be flawed?
    
   \quad *   Edge Cases \& Scenarios: Did the solution overlook critical edge cases, boundary conditions, or counter-examples?
    
   \quad *   Examples: "The argument assumes user input will always be a positive integer, failing to account for negative numbers or zero.", "The conclusion that A causes B lacks a clear, causal link."\\

2. Knowledge Depth \& Breadth:

   \quad *   Domain-Specific Understanding: Is the use and interpretation of key technical terms or domain-specific concepts accurate and sufficiently deep?
    
   \quad *   Authoritative Sourcing: Could the argument be strengthened by referencing more authoritative, credible, or up-to-date sources?
    
   \quad *   Multifaceted Perspectives: Could the problem be approached from different angles (e.g., historical, economic, technological) to yield a more comprehensive insight?
    
   \quad *   Examples: "The analysis of 'disruptive innovation' is superficial and doesn't engage with Christensen's core theory.", "Citing recent academic papers or industry reports would lend more weight to the conclusion."\\

3. Strategy \& Structure:

   \quad *   Problem Decomposition: Could the problem be broken down into smaller, more manageable sub-problems more effectively? Is the current approach to decomposition optimal?
    
   \quad *   Frameworks \& Models: Would applying a formal analytical framework or mental model (e.g., SWOT, First-Principles Thinking, MECE) lead to a more robust or structured answer?
    
   \quad *   Structural Clarity: Is the overall structure of the answer logical and easy to follow? Do the paragraphs and arguments flow coherently?
    
   \quad *   Examples: "The solution is presented as a flat list of points; a 'Pyramid Principle' (Thesis-Arguments-Data) structure would be more persuasive.", "A clear, multi-dimensional evaluation rubric is missing when comparing Option A and Option B."\\

4. Precision in Expression:

\quad    *   Linguistic Ambiguity: Does the solution use vague, ambiguous, or overly subjective language where precision is required?
    
\quad    *   Clarity of Definitions: Are key concepts defined clearly and used consistently throughout the response?
    
\quad    *   Examples: "The use of words like 'might' and 'potentially' weakens the argument; it should be replaced with data-backed assertions where possible.", "The definition of 'success' shifts between paragraphs, leading to a confusing argument."\\

Output Requirements:

*   Strictly adhere to the three-part structure: "Part 1: Reasoning Trajectory Summary" and "Part 2: Final Answer" and "Part 3: Key Areas for Improvement.".

*   In Part 3, use bullet points to clearly list each suggestion for improvement.

*   Your analysis should be objective, constructive, and aimed at elevating the quality of the reasoning.
\end{promptbox}

\newpage
\begin{promptbox}{Prompt: Critic with Code Execution (Bold Content is Re-Solver variant)}{}

\#\# Problem

\{query\}\\

\#\# Student's Solution

\{solution\_summary\}\\

\textbf{Last round answer is: \{last\_round\_answer\}. Please re-answer it.}\\

\#\# Your Job
You should critically check the student's solution to the problem, then correct it if needed and write your own answer.\\

Solve the problem with the help of feedback from a code executor. Every time you write a piece of code between $<$code$>$ and $<$/code$>$, the code inside will be executed. For example, when encountering numerical operations, you might write a piece of code to interpret the math problem into python code and print the final result in the code. Based on the reasoning process and the executor feedback, you could write code to help answering the question for multiple times (either for gaining new information or verifying). There are also several integrated functions that can be used to help you solve the problem. The available functions are:

1. web\_search(keywords), this function takes keywords as input, which is a string, and the output is a string containing several web information. This function will call a web search engine to return the search results. This function is especially useful when answering knowledge-based questions.

2. web\_parse(link:str, query:str), this function takes the link and query as input, and the output is a string containing the answer to the query according to the content in this link. This function is useful when looking into detail information of a link.\\

Your workflow for solving the problem follow these steps:

- Step 1: First, analyze the question. If it can be answered directly, provide the answer immediately. If information retrieval is required to support the answer, proceed to Step 2 and Step 3.

- Step 2: Web Search \& Parse (Verification \& Detail): Use `web\_search` to find relevant web pages for verification or supplementation. If a specific link from the search results seems particularly useful, use `web\_parse` to extract detailed information from that page.

- Step 3: Evaluate and Supplement: After receiving results from 'web\_search' or 'web\_parse', evaluate them carefully. Treat this information as a supplement to your background knowledge, not as absolute truth. This supplementary context may be incomplete or require further verification.\\

- You should not be overconfident in your knowledge and reasoning.\\

- Each time you write code put the code into $<$code$>$$<$/code$>$ snippet, and the results must be printed out through print function. Please strictly follow Python's indentation rules; do not add any extra indentation to the code. Pause after submitting any code for information retrieval or scientific computation; resume analysis only once the code has finished running.\\

For example:
    
1. If you want to use the function of web\_search(keywords), will say $<$code$>$\\
keywords=...\\
results=web\_search(keywords)\\
print(results)\\
$<$/code$>$ to call the function.

2. If you want to use the function of web\_parse(link, query), will say $<$code$>$\\
link=...\\
query=...\\
results=web\_parse(link, query)\\
print(results)\\
$<$/code$>$ to call web\_parse function.

3. If you want to do computation, You will write code for accurate result: $<$code$>$\\
a = 123\\
b = 456\\
print(a+b)\\
$<$/code$>$.\\

- Put your final answer in $<$answer$>$$<$/answer$>$ with boxed.
\end{promptbox}

\newpage
\begin{promptbox}{Prompt: Selector with Code Execution (Bold Content is Re-Selector variant)}{}

You are a diligent and precise judge. You should choose the correct response from the following \{PARALLEL\_NUM\} responses to the problem. To maximize confidence and accuracy, you must rigorously verify each response using tool-based searches (`web\_search` and `web\_parse`), with a focus on precision and critical evaluation of sources.\\

The problem is: 
\{query\}\\

The responses are:
\{responses\}\\

\textbf{Based on historical selections and their entropy values, re-perform the selection to improve the confidence and accuracy of the model's selection.
\{last\_selection\}}\\

\#\# Your Task

You should thoroughly analyse each response carefully by writing codes and choose the most correct one from \{PARALLEL\_NUM\} responses. Every time you write a piece of code between $<$code$>$ and $<$/code$>$, the code inside will be executed. For example, when encountering numerical operations, you might write a piece of code to interpret the math problem into python code and print the final result in the code. Based on the reasoning process and the executor feedback, you could write code to help answering the question for multiple times (either for gaining new information or verifying). There are also several integrated functions that can be used to help you solve the problem. The available functions are:

1. web\_search(keywords), this function takes keywords as input, which is a string, and the output is a string containing several web information. This function will call a web search engine to return the search results. This function is especially useful when answering knowledge-based questions.

2. web\_parse(link:str, query:str), this function takes the link and query as input, and the output is a string containing the answer to the query according to the content in this link. This function is useful when looking into detail information of a link.\\

\#\# Your Task Process is as Follows:

\#\#\# 1. Preliminary Analysis and Search Planning (Plan)

- {Analyze the Core of the Problem}: First, what is the essence of the problem? Which key concepts, facts, or logical relationships are involved?

- {Identify Knowledge Gaps}: To answer this question correctly, what key information do you need to verify or obtain? Which statements in the options may be ambiguous or require fact-checking?

- {Formulate a Search Strategy}: For each key point and the options that need verification, what kind of {keywords} should you use for `web\_search`? Please list the initial list of search keywords.\\

\#\#\# 2. Execute Iterative Search and In-depth Analysis (Search \& Parse)

- {First-round Search}: Use the keywords you consider most core for `web\_search` to obtain background knowledge and an overview of the problem.

- {Evaluation and Deepening}: Browse the search results and identify authoritative and relevant information sources (such as encyclopedias, official documents, academic articles, and well-known technology websites). Use the `web\_parse` tool to extract detailed information directly related to the problem from these high-quality links.

- {Targeted Verification}: Conduct {targeted searches and analysis} for each option. For example, for Option A, you can search for "Is the core claim in Option A valid?" or "The correct definition of the concept in Option A". Repeat this process for Options B, C, and D. Pay special attention to options that are contradictory or expressed in absolute terms.

- {Cross-verification}: Do not rely on a single information source. For key assertions, try to conduct search verification from another independent source (e.g., a different website or media outlet) to see if there is consensus or disagreement.\\

\#\#\# 3. Comprehensive Comparison and Reasoning (Synthesize \& Reason)

- {Information Organization}: Based on the collected information, briefly summarize the supporting and opposing evidence related to each candidate answer.

- {Logical Reasoning}: Conduct logical reasoning combined with verified facts. Even if a candidate answer "sounds" reasonable, is it inconsistent with verified facts or basic logic?

- {Identify Traps}: Reflect on whether any candidate answer takes advantage of common misunderstandings or outdated information. Does the evidence you found refute these traps?\\

\#\#\# 4. Provide Final Judgment and Evidence (Conclude)

- {Final Selection}: What is your final judgment on which candidate answer is correct? Please answer clearly.

- {Evidence Statement}: Clearly and concisely state the \textbf{core evidence} for your judgment, and cite credible sources from `web\_parse` as much as possible. Explain why this candidate answer is the most compelling and why the other candidate answers are excluded.\\

\#\# Tool Usage Requirements:

- After each use of `web\_search`, evaluate the relevance and authority of the results.

- Prioritize using `web\_parse` to obtain accurate information from high-authority, high-relevance links, rather than relying solely on search summaries.

- Your thinking process should fully demonstrate the above steps.

- You should not be overconfident in your knowledge or reasoning.

- Each time you write code put the code into $<$code$>$$<$/code$>$ snippet, and the results must be printed out through print function. Please strictly follow Python's indentation rules; do not add any extra indentation to the code. Pause after submitting any code for information retrieval or scientific computation; resume analysis only once the code has finished running.\\

For example:

1. If you want to use the function of web\_search(keywords), will say $<$code$>$\\
keywords=...\\
results=web\_search(keywords)\\
print(results)\\
$<$/code$>$ to call the function.

2. If you want to use the function of web\_parse(link, query), will say $<$code$>$\\
link=...\\
query=...\\
results=web\_parse(link, query)\\
print(results)\\
$<$/code$>$ to call web\_parse function.

3. If you want to do computation, You will write code for accurate result: $<$code$>$\\
a = 123\\
b = 456\\
print(a+b)\\
$<$/code$>$.\\
- Finally, you should analyze whether each response is correct.\\

{Notice}

1. Do not trust the information, reference or any assumptions in the response easily. You must write codes to verify it before reaching a conclusion.

2. Do not be influenced by the majority number of final answers. They may collude to deceive you!

3. The return of web functions may be empty due to network issue, you can try it again.

4. You should collect enough information from web functions to verify each response.\\

\#\# Format Requirement\\
Your response MUST follow this exact format:\\

VERIFICATION:\\
$[$ Your detailed verification process for response 1 here $]$\\
$[$ Your detailed verification process for response 2 here $]$\\
$...$\\
$[$ Your detailed verification process for response \{PARALLEL\_NUM\} here $]$\\

CROSS VERIFICATION\\
$[$ Search for multiple perspectives on contentious points to reduce AI hallucinations $]$

CONCLUSION:\\
$[$ Your brief summarization of the verification process and the final decision $]$\\

FINAL DECISION: $<$select$>$Response X$<$/select$>$

Replace X with the response index, for example 1, 2, ..., up to \{PARALLEL\_NUM\}. The $<$select$>$ tags are required.
\end{promptbox}

\section{Limitations and Future Work}
Despite its performance gains, ReThinker has several limitations.
First, the sequential Solver–Critic–Selector pipeline introduces additional latency, increasing wall-clock time by approximately \textbf{1.5$\times$} compared to single-pass baselines.
Although uncertainty-aware gating reduces unnecessary computation, the inherently sequential structure remains a bottleneck.

Second, the current \textbf{128K} context window is insufficient for the most challenging long-horizon scientific tasks, motivating future exploration of extended context lengths or more advanced context management strategies.

Third, ReThinker currently relies on generic search and validation tools.
Integrating specialized tools—such as symbolic theorem provers, property predictors, or structured database query engines—could further improve performance on domains such as mathematics and chemistry.
However, this would require addressing tool-specific failure modes and adaptive invocation strategies.

Overall, these limitations highlight that ReThinker’s primary strength—adaptive orchestration—does not fully offset the latency and context constraints inherent to multi-stage reasoning pipelines.
Future work should focus on parallelizing stage execution, developing dynamic context management mechanisms, and enabling tool-augmented reflection that can reliably invoke specialized reasoning modules beyond generic search.

\end{document}